\documentclass[lettersize,journal]{IEEEtran}
\usepackage{amsmath,amsfonts}
\usepackage{algorithmic}
\usepackage{algorithm}
\usepackage{array}
\usepackage[caption=false,font=normalsize,labelfont=sf,textfont=sf]{subfig}
\usepackage{textcomp}
\usepackage{stfloats}
\usepackage{url}
\usepackage{verbatim}
\usepackage{graphicx}
\usepackage{cite}
\usepackage{booktabs}
\usepackage{multirow}
\usepackage{xcolor}
\usepackage{tabularx}

\definecolor{best_red}{RGB}{192, 0, 0}
\definecolor{second_blue}{RGB}{0, 102, 204}

\newcolumntype{C}{>{\centering\arraybackslash}X}

\usepackage{colortbl}
\usepackage{xcolor}

\definecolor{metricbg}{HTML}{E8EEFA} % 浅蓝（IoU/Prec/Rec）
\definecolor{f1bg}{HTML}{F6E7D8}    % 浅米（F1）

\usepackage{hyperref}

\usepackage{hyperref}

\usepackage{makecell}

\usepackage{orcidlink}

\begin{document}

% \title{DDA-OVCD: Dual Domain Aware Open Vocabulary Change Detection for remote sensing Imagery.}

\title{ReA-OVCD: Training-Free Open-Vocabulary Change Detection \\ via Semantic-Spatial Reliability Assessment}

% \title{ReA-OVCD: Reliability-Aware Open-Vocabulary Change Detection via Semantic and Spatial Refinement}

% \author{IEEE Publication Technology,~\IEEEmembership{Staff,~IEEE,}
%         % <-this % stops a space
% \thanks{This paper was produced by the IEEE Publication Technology Group. They are in Piscataway, NJ.}% <-this % stops a space
% \thanks{Manuscript received April 19, 2021; revised August 16, 2021.}}

\author{
Hongming Zhu\orcidlink{0000-0001-5795-5279},
Huaji Chen\orcidlink{0009-0005-2198-5325},
Bowen Du\orcidlink{0000-0002-3755-4870},~\IEEEmembership{Member,~IEEE,}
Sicong Liu\orcidlink{0000-0003-1612-4844},~\IEEEmembership{Senior Member,~IEEE,}
and Qin Liu\orcidlink{0000-0002-9352-1694}

\thanks{
This work was supported in part by the National Key R\&D Program of China under Grant 2023YFC3805305, in part by the Program of Shanghai Subject Chief Scientist under Grant 23XD1433900, in part by the National Natural Science Foundation of China under Grant 62576250, and in part by the National Natural Science Foundation of China under Grant 62406227. (Corresponding author: Bowen Du.)
}

\thanks{
Hongming Zhu, Huaji Chen, Bowen Du, and Qin Liu are with the School of Computer Science and Technology, Tongji University, Shanghai 200092, China
(e-mail: zhu\_hongming@tongji.edu.cn;
2534072@tongji.edu.cn;
bowendu@tongji.edu.cn;
qin.liu@tongji.edu.cn).
}
\thanks{
Sicong Liu is with the College of Surveying and Geo-Informatics, Tongji University, Shanghai 200092, China
(e-mail: sicong.liu@tongji.edu.cn).
}
}

% The paper headers
\markboth{Journal of \LaTeX\ Class Files,~Vol.~14, No.~8, August~2021}%
{Shell \MakeLowercase{\textit{et al.}}: A Sample Article Using IEEEtran.cls for IEEE Journals}

% \IEEEpubid{0000--0000/00\$00.00~\copyright~2021 IEEE}
% Remember, if you use this you must call \IEEEpubidadjcol in the second
% column for its text to clear the IEEEpubid mark.

\maketitle

\begin{abstract}
Unlike traditional remote sensing change detection that relies on predefined categories, Open-Vocabulary Change Detection (OVCD) identifies land cover changes flexibly using arbitrary text prompts. 
% However, existing methods suffer from an inherent trade-off when modeling changes: instance-level comparison overlooks fine-grained semantic variations (e.g., partial building extensions), while direct pixel comparison proves unreliable, yielding unstable responses and boundary artifacts due to semantic ambiguity and spatial inconsistency. 
However, most existing OVCD methods rely on instance-level matching for stable correspondence but may overlook fine-grained variations (e.g., partial building extensions). Dense pixel-level comparison is more flexible, yet direct semantic comparison often produces unreliable candidate changes due to semantic ambiguity and spatial inconsistency.
To this end, we propose ReA-OVCD, an efficient training-free framework that revisits pixel-level OVCD from a reliability assessment perspective. It first derives candidate change regions from pixel-wise semantic discrepancies to retain flexible localization. Instead of directly trusting these candidates, ReA-OVCD applies a two-stage semantic-spatial reliability assessment. The semantic stage evaluates whether a label discrepancy is supported by meaningful distributional and response-level changes, while the spatial stage validates whether a candidate region contains stable interior evidence rather than only boundary-induced responses.
Extensive experiments across LEVIR-CD, WHU-CD, DSIFN, and SECOND show that the proposed framework improves the reliability of pixel-level OVCD and consistently outperforms state-of-the-art approaches, achieving $\mathrm{F}_{1}^{C}$ improvements of 3.54\% to 8.45\% while maintaining superior computational efficiency.
The code is available at \href{https://github.com/Funny0101/ReA-OVCD}{https://github.com/Funny0101/ReA-OVCD}.
\end{abstract}


\section{Introduction}

\IEEEPARstart{A}{ccurate} spatiotemporal monitoring is crucial for understanding Earth's surface dynamics. 
In this context, remote sensing change detection serves as a foundational tool, aiming to identify changes in land cover over time from a pair of images captured in the same geographical area\cite{rs16132355}. It plays a vital role in climate change tracking, disaster damage assessment, and urban expansion monitoring\cite{Song2018,ZHENG2021112636,9955391}.

Depending on the semantic level of the task, traditional change detection can be categorized into binary change detection (BCD) and semantic change detection (SCD). The former focuses on determining whether changes have occurred yet lacks semantic understanding~\cite{changeformer,samcd}; the latter further introduces class label information, enabling the identification of specific land cover change types~\cite{hrscd,scannet}. However, both paradigms are typically based on the closed-set assumption, where the category space is predefined and shared between training and inference. In practice, land cover types are diverse and researchers often need to flexibly define targets of interest according to specific tasks (e.g., composite concepts like \textit{impervious surface} rather than low-level classes such as \textit{pavement} that are provided during training~\cite{WENG201234}). This mismatch significantly limits the adaptability of conventional methods when encountering unseen categories.
In recent years, breakthroughs in foundational models \cite{sam,clip,dino} have made open-vocabulary change detection (OVCD) increasingly feasible. OVCD aims to identify semantic changes using natural language descriptions without the need for predefined category labels. 
To better preserve this flexibility, training-free OVCD has emerged as an appealing paradigm because it avoids introducing category bias and reduces the risk of overfitting to a fixed label space.

Existing methods often rely on instance-level matching\cite{dynamicearth,omniovcd}, which provides relatively stable comparisons but may overlook localized semantic changes within an object (e.g., partial building extensions). In contrast, pixel-level methods~\cite{univcd,seg2change} offer a more fine-grained alternative by comparing predictions at each location, yet are prone to introducing unreliable responses.
Specifically, due to the open category space, pixel predictions are often unstable and exhibit ambiguous distributions, with no clearly dominant class. Directly comparing discretized labels can therefore convert small response perturbations near decision boundaries into abrupt label changes, producing false positives even when the underlying semantic responses remain nearly identical.
Moreover, slight geometric misalignments and segmentation uncertainties near object boundaries can introduce systematic boundary shifts across temporal predictions. Treating each pixel independently and ignoring spatial context amplifies these shifts and produces strip-like spurious change responses along object edges.

To address these issues, we propose ReA-OVCD, an efficient training-free framework that revisits pixel-level OVCD via semantic-spatial reliability assessment. Specifically, ReA-OVCD first employs a unified open-vocabulary perceiver to efficiently encode semantic information across bi-temporal images. Subsequently, it uses discrete label discrepancies only for candidate change generation to preserve flexible localization, and then assesses their reliability from semantic and spatial domains rather than directly accepting them as final changes:

For semantic inconsistency induced by ambiguous open-vocabulary responses, our hypothesis is that reliable changes are expected to exhibit both structured distributional shifts across categories and significant response-level changes, whereas incidental label flips caused by minor response perturbations usually lack such evidence. Based on this, we introduce Semantic Change Reasoning (SCR) to filter semantically unreliable candidates by combining distribution-level divergence and response variation while preserving semantically meaningful ones.
For spatial inconsistency caused by segmentation uncertainty and boundary misalignment, our key observation is that a reliable changed region should contain stable interior support rather than being concentrated only around object boundaries. To this end, we design a Boundary-aware Change Refinement (BCR) module that evaluates the reliability of candidate change regions by incorporating spatial context. In practice, this corresponds to retaining regions supported by enough reliable interior pixels while removing the artifacts dominated by boundary-induced responses.

Overall, ReA-OVCD decouples dense OVCD into pixel-level candidate generation and semantic-spatial reliability assessment, where semantic response evidence and region-level interior support are jointly used to filter unreliable candidates caused by ambiguous responses and boundary inconsistency. This design preserves the flexibility of pixel-level change perception without relying on instance-level matching or task-specific adaptation, keeping the overall framework training-free and computationally efficient.
In summary, the main contributions of this paper are as follows:
\begin{itemize}
    \item We propose ReA-OVCD, an efficient training-free OVCD framework that enables reliable change detection while preserving fine-grained change perception through reducing false positives introduced by direct pixel-wise comparison in open-vocabulary settings.
    \item We develop a semantic-spatial reliability assessment mechanism for candidate change filtering: SCR mitigates unreliable changes caused by semantic ambiguity; BCR reduces artifacts near boundaries that arise from geometric misalignment and segmentation uncertainties.
    \item Extensive experiments on multiple datasets demonstrate that the proposed method significantly outperforms existing approaches, validating its effectiveness, efficiency, and generalization capability in OVCD tasks.
\end{itemize}

\section{Related Work}
\subsection{Segment Anything Model}
SAM\cite{sam} pioneered the promptable universal image segmentation paradigm, enabling unified zero-shot promptable segmentation capabilities across multiple downstream visual tasks through pretraining on large-scale datasets. Subsequently, numerous models emerged, focusing on enhancing its performance and efficiency. For instance, HQ-SAM\cite{hqsam} introduced high-quality tokens to enhance segmentation quality; methods like FastSAM\cite{fastsam}, MobileSAM\cite{mobilesam}, TinySAM\cite{tinysam}, and EfficientViT-SAM\cite{efficientsam} significantly reduced model inference costs through lightweight network architectures or distillation strategies, enabling real-time inference on resource-constrained devices. Building upon this foundation, the SAM series gradually expanded to more complex visual tasks. SAM 2\cite{sam2} extended segmentation capabilities from single-frame images to video scenes, enabling cross-frame object tracking and mask propagation through the introduction of memory attention mechanisms. The newly proposed SAM 3\cite{sam3} further introduces the Promptable Concept Segmentation task, which can directly localize and segment arbitrary semantic concepts. This capability provides the semantic parsing basis for open-vocabulary change detection.

\subsection{Traditional Change Detection}
Change detection aims to identify land cover changes by comparing feature differences between dual-phase images. Traditional approaches predominantly employ CNN-based Siamese networks\cite{siamesenet,snunetcd,tinycd} to extract multi-temporal features, combined with attention-based feature fusion modules to enhance change detection capabilities. However, constrained by sparse annotated data\cite{zeroscd,rs12101688}, these models exhibit limited understanding of change semantics. With advancements in foundational models, studies like ChangeCLIP\cite{changeclip}, TextSCD\cite{textscd}, and CKCD\cite{ckcd} leverage CLIP's visual-text alignment capability to introduce textual modalities, thereby enhancing semantic representation and suppressing false changes. Meanwhile, SAM-based change detection research continues to emerge: SAMCD\cite{samcd} and SCM\cite{scm} utilize SAM's robust segmentation and localization capabilities to identify change regions, while AnyChange\cite{anychange} achieves zero-shot change localization through latent space matching. Furthermore, approaches like BiSAM-CD\cite{bisamcd} and Change3D\cite{change3d} redefine change detection from a video modeling perspective. Additionally, studies like UniChange\cite{unichange} attempt to unify BCD and SCD tasks using multimodal large language models. However, these approaches still face significant limitations: while some can detect changes under zero-shot conditions, they cannot identify change categories; methods capable of semantic recognition typically rely on fixed category sets or specific dataset training, making them unsuitable for practical application scenarios. Therefore, it is necessary to explore change detection beyond predefined label spaces, where target categories can be specified more flexibly according to user needs.

\subsection{Open-Vocabulary Change Detection}
OVCD aims to identify semantic changes without the need for predefined categories. Existing methods can be broadly categorized into training-based and training-free approaches.
Training-based approaches typically construct dedicated datasets and training strategies tailored to the change detection task. Semantic-CD~\cite{semanticcd} transforms the problem as a comparison of bi-temporal semantic predictions, ViLaCD-R1~\cite{vilacdr1} utilizes vision-language alignment to identify change regions, UniVCD~\cite{univcd} fuses features from SAM and CLIP through feature alignment, and Seg2Change~\cite{seg2change} allows spatial context to be learned for change prediction through a trained category-agnostic change adapter. While these methods improve task-specific discriminative capability, their performance depends on the distribution of training data and may compromise open-vocabulary generalization to some extent.
In contrast, training-free methods are more aligned with the open-vocabulary setting because they avoid additional category-specific adaptation. DynamicEarth~\cite{dynamicearth} and OV-CD~\cite{ovcd} perform change detection through loosely coupled multi-model cascades, where separate modules handle different sub-tasks, but such designs may suffer from feature inconsistency and error accumulation. AdaptOVCD~\cite{adaptovcd} enhances change representation through multi-level feature fusion, while OpenDPR~\cite{opendpr} leverages a diffusion model to generate change-aware examples that guide detection. However, these approaches still rely on multi-stage or multi-model designs, resulting in increased architectural complexity and reduced structural coherence. 
Recent OVSS studies such as SegEarth-OV3~\cite{segearthov3} show that unified foundation models can provide dense pixel-level semantic responses for arbitrary text queries in a training-free manner, offering a promising semantic basis for OVCD. Built upon this capability, OmniOVCD~\cite{omniovcd} streamlines OVCD with SAM-3-based semantic parsing and cross-instance matching, but its instance-level comparison may overlook localized semantic variations within an object. To preserve such localized change perception, dense pixel-level comparison provides a more flexible alternative. However, directly comparing pixel-wise semantic predictions in a training-free setting can produce unreliable changes due to ambiguous open-vocabulary responses and spatial inconsistency around object boundaries. ReA-OVCD addresses this problem by revisiting pixel-level OVCD from a semantic-spatial reliability assessment perspective, without additional training.

\begin{figure*}[!t]
\centering
\includegraphics[width=\textwidth]{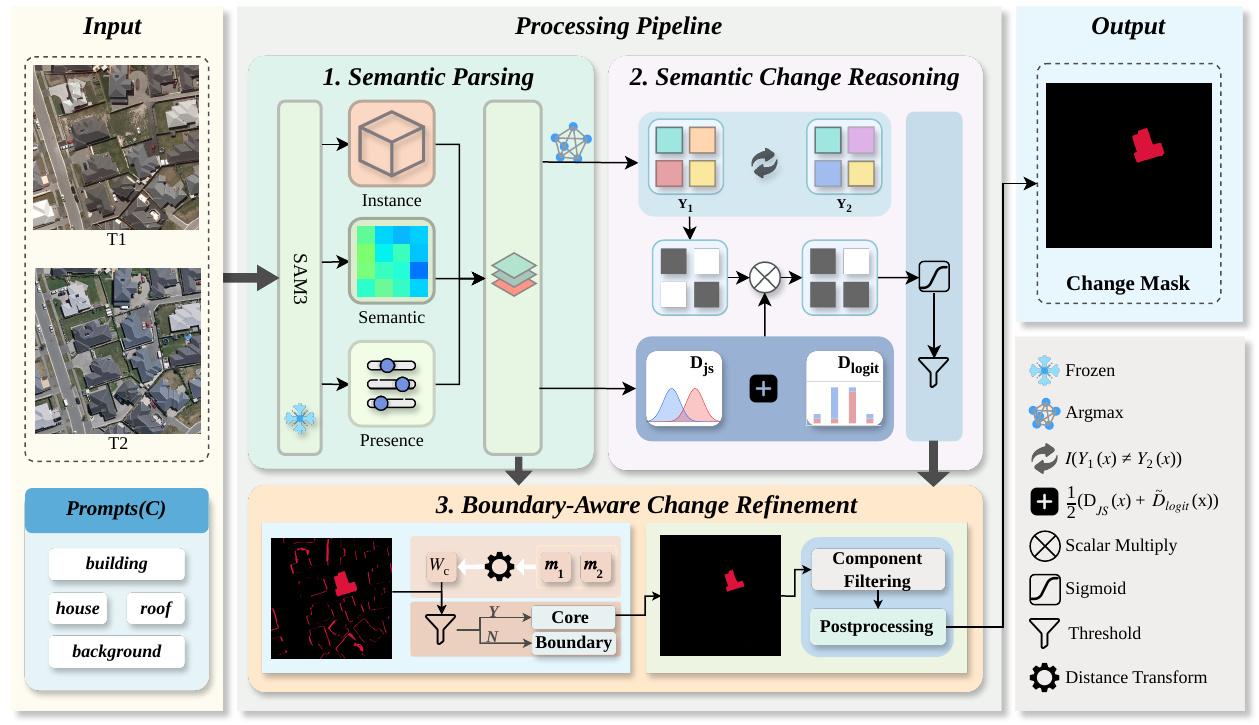}
\caption{Overall architecture of the proposed ReA-OVCD framework. Given bi-temporal images T1 and T2 and a set of open-vocabulary queries C, we first employ a frozen SAM-3 to extract and fuse multi-head outputs, producing pixel-wise semantic logits and corresponding label maps. Initial candidate changes are derived from label discrepancies. The SCR stage removes semantically unreliable candidates using Jensen-Shannon divergence and response difference as reliability cues. The BCR stage further removes boundary-dominated candidates through distance-based confidence and connected-component validation, yielding more reliable change masks.}
\label{fig:architecture_draft}
\end{figure*}

\section{Methodology}

\subsection{Framework Overview}
Given a pair of registered remote sensing images captured at different times, denoted respectively as $I_{1}, I_2 \in \mathbb{R}^{H \times W \times 3}$, and a set of open categories $\mathcal{C} = \{c_0, c_1, \dots, c_{N-1}\}$ consisting of natural language descriptions, the objective of OVCD is to extract changes in relevant regions based on $\mathcal{C}$ without the need for additional fine-tuning for the target categories. Here, $H$ and $W$ denote the resolution of the input images, $c_0$ represents the background category, and $N$ denotes the total
number of queried classes.
Specifically, for each query $c_i$ in the set $\mathcal{C}$, a binary change mask $M_{change}^{c_i} \in \{0, 1\}^{H \times W}$ is generated independently, where 1 indicates that the region associated with that concept has changed, and 0 indicates that no change has occurred. Unlike traditional semantic change detection, the categories in OVCD are no longer restricted to a predefined label space, but are dynamically specified by arbitrary text queries.

% changed

% To address the instability of direct pixel-wise semantic comparison in open-vocabulary settings, we propose ReA-OVCD, an efficient training-free framework that follows a candidate-generation-and-filtering paradigm for dense pixel-level OVCD.
To address the instability of direct pixel-wise semantic comparison in open-vocabulary settings, we propose ReA-OVCD, an efficient training-free framework that revisits dense pixel-level OVCD through semantic-spatial reliability assessment.
As shown in Fig.~\ref{fig:architecture_draft}, the overall process consists of three stages.
First, ReA-OVCD employs a unified SAM-3-based open-vocabulary perceiver to generate dense semantic logits and label maps for bi-temporal images through multi-head fusion, avoiding redundant feature extraction in multi-model parsing pipelines.
Subsequently, discrete label discrepancies are used only to produce initial candidate change regions. SCR then evaluates whether these candidates are supported by distribution-level divergence and response-level variation, filtering semantically unreliable candidates caused by ambiguous open-vocabulary responses.
Finally, BCR further examines candidate regions from the spatial perspective. Instead of treating all candidate pixels independently, it retains regions with sufficient interior support and removes boundary-dominated candidates caused by segmentation uncertainty or slight geometric misalignment.

% origin
% To address the limitations in capturing fine-grained semantic changes while maintaining prediction reliability, we propose ReA-OVCD, an efficient training-free framework that follows a candidate-generation-and-filtering paradigm for dense pixel-level OVCD. 
% As shown in the figure~\ref{fig:architecture_draft}, the overall process primarily comprises three core stages.
% The first stage is open-vocabulary semantic feature extraction. We utilize a unified SAM-3 to generate dense semantic logits and probability distributions for the queried categories through multi-head fusion, thereby avoiding the redundant feature extraction overhead introduced by existing multi-stage parsing pipelines.
% Subsequently, starting from candidate change regions derived from discrete label discrepancies, a lightweight Semantic Change Reasoning (SCR) is performed within the semantic domain. By modeling differences in semantic distributions and changes in response intensity across temporal features, it successfully suppresses spurious changes caused by incidental label inconsistencies and retains regions with reliable semantic shifts.
% Finally, Boundary-aware Change Refinement (BCR) is performed in the spatial domain. Specifically, within sparse candidate change regions, regions lacking sufficient reliable interior pixels are removed, while those with consistent interior support are preserved, thereby improving the structural integrity of detected changes.

\subsection{Bi-temporal Open-Vocabulary Semantic Parsing}
To obtain high-quality open-vocabulary semantic features without training, we employ a frozen SAM-3 model as a unified feature extractor, which avoids the error accumulation of traditional multi-model stacking and improves computational efficiency\cite{omniovcd}.
To accommodate both the precise boundary delineation of structured objects (e.g., buildings) and the semantic spatial continuity of land cover areas (e.g., water), we draw on the design philosophy of SegEarth-OV-3\cite{segearthov3} to simultaneously activate and aggregate the outputs of both the instance and semantic heads during decoding. 

Specifically, the first step of this process is the instance aggregation operation. 
For a given query in $\mathcal{C}$, the instance heade generates $P$ candidate instances. Let their corresponding instance-level logit maps and confidence scores be denoted as $L_{inst}^{(p)} \in \mathbb{R}^{H \times W}$ and $s_{conf}^{(p)} \in [0, 1]$, respectively.
To integrate the sparse instance predictions into a continuous semantic response, we take the maximum of the weighted instance responses at the pixel level $x$:
\begin{equation}
L_{agg}^{inst}(x) = \max_{p=1}^{P} \left( L_{inst}^{(p)}(x) \cdot s_{conf}^{(p)} \right)
\end{equation}
where $L_{agg}^{inst} \in \mathbb{R}^{H \times W}$ is the aggregated instance response map. This operation suppresses interference from low-quality candidates and retains high-confidence instance boundaries. 

Subsequently, the aggregated instance response is fused with the dense semantic logit map $L_{sem} \in \mathbb{R}^{H \times W}$ produced by the semantic head via pixel-wise max-pooling:
\begin{equation}
L_{fused}(x) = \max \left( L_{sem}(x), L_{agg}^{inst}(x) \right),
\end{equation}
This fusion strategy, yielding $L_{fused} \in \mathbb{R}^{H \times W}$, combines instance-level precision with semantic-level global consistency, thereby enhancing the overall modeling capability for targets at different scales.

Furthermore, in open-vocabulary scenarios, a large-scale query vocabulary introduces a significant number of categories that do not actually exist in the image, leading to noise and hallucination issues in semantic responses. To address this, we employ the model's category presence score $S_{pres} \in [0, 1]$ to perform soft suppression on the fused response:
\begin{equation}
L_{final}(x) = L_{fused}(x) \cdot S_{pres}
\end{equation}
This soft suppression operation yields $L_{final} \in \mathbb{R}^{H \times W}$, which explicitly reduces the response strength of irrelevant categories and improves the reliability of semantic predictions.

Given that a single semantic category may correspond to multiple synonymous query terms, we let $\mathcal{Q}_c \subset \mathcal{C}$ denote the subset of queries corresponding to category $c$, and perform maximum aggregation over these synonymous queries at the category level to obtain the category-specific response map:
\begin{equation}
L^{c}(x) = \max_{q \in \mathcal{Q}_c} L_{final, q}(x),
\end{equation}
where $L^{c} \in \mathbb{R}^{H \times W}$. By stacking the responses of all $N$ unique semantic categories, we obtain the holistic semantic logit tensor $\mathbf{L} \in \mathbb{R}^{N \times H \times W}$. 
Finally, we extract the initial class label $Y(x)$ corresponding to the maximum response value across the category dimension for each pixel:
\begin{equation}
Y(x) = \arg\max_{c \in \mathcal{C}} L^c(x)
\end{equation}
Concurrently, to filter out background predictions with low confidence, we introduce a probability threshold $\theta_{prob}$. If a pixel's maximum logit response falls below this threshold, it is forcibly assigned to the background class $c_0$. By applying the aforementioned processing pipeline to the dual-phase images separately, we obtain their respective segmentation logit tensors $\mathbf{L}_1, \mathbf{L}_2 \in \mathbb{R}^{N \times H \times W}$, along with the filtered semantic label maps $Y_1, Y_2 \in \mathbb{R}^{H \times W}$, which are subsequently utilized by the following procedures.

\subsection{Semantic Change Reasoning}

Existing change detection methods typically rely on comparing cross-temporal predictions after discretizing continuous semantic responses using \textit{argmax}. However, this strategy implicitly assumes that semantic predictions are stable, which does not hold in open-vocabulary settings. Due to the ambiguity of category boundaries, pixel-level predictions often exhibit uncertain distributions, where minor perturbations can lead to label flips and consequently introduce spurious changes. 
% To filter such unstable label-discrepancy candidates, we introduce the SCR module, which keeps discrete label mismatch as a candidate-generation cue and examines whether each candidate is supported by continuous cross-temporal semantic evidence.}
To filter such unstable label-discrepancy responses, we introduce the SCR module, which uses discrete label mismatch only as an initial change indication and examines whether it is supported by continuous cross-temporal semantic evidence.

% To address this, we propose Semantic Change Reasoning (SCR) module, which reassesses the reliability of candidate changes by modeling cross-temporal differences in continuous semantic distributions instead of relying solely on discrete label comparison.

First, based on the discrete prediction results $Y_1$ and $Y_2$ obtained above, we construct a preliminary hard change indicator map $H_{change} \in \{0, 1\}^{H \times W}$:
\begin{equation}
H_{change}(x) = \mathbb{I} \big( Y_1(x) \neq Y_2(x) \big)
\end{equation}
where $\mathbb{I}(\cdot)$ is the standard indicator function, which evaluates to 1 if the condition is met and 0 otherwise. 
While this pixel-wise operation is significantly more efficient than computationally expensive cross-instance matching for capturing potential change regions, it still contains a large number of spurious changes due to label instability. Therefore, further constraints and filtering are required within the continuous semantic space.

Based on the dense logit outputs $L_1$ and $L_2$ from the dual-temporal branches, we first convert them into normalized probability distributions along the class dimension, which enables a meaningful comparison of semantic distributions. Specifically, we apply the \textit{Softmax} function to obtain the pixel-level class probabilities:
\begin{equation}
P_t^c(x) = \frac{\exp(L_t^c(x))}{\sum_{j \in \mathcal{C}} \exp(L_t^j(x))}, \quad t \in \{1,2\}
\end{equation}
This formulation produces the continuous probability distribution $P_t \in [0, 1]^{N \times H \times W}$. Since reliable semantic changes typically involve substantial redistribution of semantic probabilities across categories, we employ the Jensen-Shannon (JS) divergence to measure the overall difference between the two semantic distributions.
Compared to simple subtraction, it provides a symmetric and robust measure of the shape discrepancy between two probability distributions. It is formulated as:
% \begin{equation}
% M^c(x) = \frac{1}{2} \big(P_1(x) + P_2(x)\big),
% \end{equation}
\begin{equation}
M^c(x)=\frac{1}{2}\left(P_1^c(x)+P_2^c(x)\right),
\end{equation}
\begin{equation}
\begin{aligned}
D_{JS}(x) &= \frac{1}{2} \sum_{c=1}^{N} P_1^c(x) \log \left( \frac{P_1^c(x)}{M^c(x)} \right) \\
&\quad + \frac{1}{2} \sum_{c=1}^{N} P_2^c(x) \log \left( \frac{P_2^c(x)}{M^c(x)} \right),
\end{aligned}
\end{equation}
where $M^c(x)$ is the mean distribution and $D_{JS} \in \mathbb{R}^{H \times W}$ is the pixel-wise JS divergence. 
While effective, it may not fully characterize variations in response magnitude. Therefore, to additionally capture confidence-level response changes for highly activated categories, we further calculate the maximum response difference $D_{logit} \in \mathbb{R}^{H \times W}$ at the raw logit level:

\begin{equation}
D_{logit}(x) = \max_{c \in \{1, \dots, N\}} \left| L_1^c(x) - L_2^c(x) \right|,
\end{equation}
Subsequently, we perform a weighted fusion of these two complementary metrics to formulate a comprehensive continuous change representation:
\begin{equation}
S_{fusion}(x) = \frac{1}{2} D_{JS}(x) + \frac{1}{2}\,\tilde{D}_{logit}(x),
\end{equation}
where $\tilde{D}_{logit}$ denotes the min-max normalized response difference, which is used to align its numerical scale with $D_{JS}$ for stable and balanced fusion. Finally, by masking this continuous fusion score with the initial hard change constraint, we derive the reliable semantic change score:
\begin{equation}
S_{change}(x) = H_{change}(x) \cdot S_{fusion}(x),
\end{equation}
To ensure a consistent score range, the fused change score is normalized over the entire image pair via min-max normalization, yielding $\widetilde{S}_{\mathrm{change}}(x)$. The initial binary change mask is then obtained as:
\begin{equation}
M_{\mathrm{base}}(x)=
\mathbb{I}\!\left(
\widetilde{S}_{\mathrm{change}}(x)>\theta_{\mathrm{change}}
\right).
\end{equation}

% By setting a threshold $\theta_{change}$, we obtain the refined base change mask $M_{base} \in \{0, 1\}^{H \times W}$:
% \begin{equation}
% M_{base}(x) = \mathbb{I} \big(S_{change}(x) > \theta_{change} \big).
% \end{equation}

% % changed
% This formulation also provides an efficient way to unify BCD and SCD within the same open-vocabulary pipeline. Instead of treating SCD as multiple independent binary change detection tasks like current mainstream methods do, we obtain the class-specific candidate mask for a queried class $c$ by combining the shared base change mask $M_{base}$ with the discrepancy between the binary class-membership masks of $m_1^c$ and $m_2^c$:}

% origin
This formulation also provides an efficient way to unify BCD and SCD within the same open-vocabulary pipeline. Instead of treating SCD as multiple independent binary change detection tasks as current mainstream methods do, we formulate the change detection process as a category-specific query task.
Specifically, for a queried target class $c$, we first utilize the dual-temporal discrete labels $Y_1$ and $Y_2$ to generate its independent binary masks: $m_{1}^c(x) = \mathbb{I}(Y_1(x) = c)$ and $m_{2}^c(x) = \mathbb{I}(Y_2(x) = c)$. By combining these masks with the global indicator $M_{base}$, we define the initial change prediction for class $c$ as:
\begin{equation}
M_{base}^c(x) = M_{base}(x) \cdot \mathbb{I} \big( m_{1}^c(x) \neq m_{2}^c(x) \big)
\end{equation}

Notably, the entire process relies solely on lightweight element-wise tensor operations. Without introducing any additional complex procedures, its computational overhead is negligible compared with the backbone inference process.

\subsection{Boundary-aware Change Refinement}
% origin
% Although the SCR module effectively mitigates spurious changes caused by semantic uncertainty, the candidate change mask $M_{base}^c$ lacks explicit spatial structural constraints. By treating each pixel as an independent entity, pixel-wise comparisons inherently overlook spatial continuity. Consequently, due to the coupled effects of segmentation uncertainty and slight geometric misalignment in bi-temporal remote sensing imagery, systematic boundary shifts frequently occur. These shifts are amplified during cross-temporal pixel-wise comparison, manifesting as strip-like false positives along object edges. Since these artifacts lack stable interior support, their fundamental origin is boundary ambiguity rather than genuine semantic structural transformation. To address this, we propose a Boundary-Aware Change Refinement (BCR) module to evaluate the spatial reliability of candidate regions.

% changed
Although SCR filters semantically unreliable candidates, the resulting mask $M_{base}^c$ is still produced from pixel-wise discrepancies and therefore lacks explicit spatial support. In bi-temporal remote sensing images, segmentation uncertainty and slight geometric misalignment can lead to inconsistent boundaries across phases. When such boundary shifts are compared pixel by pixel, they often appear as strip-like candidate changes along object edges. These candidates usually lack stable interior support and are more related to boundary ambiguity than to reliable regional changes. To address this, we introduce the Boundary-aware Change Refinement (BCR) module to evaluate the spatial reliability of candidate regions.

% origin
We observe that pseudo-changes are predominantly clustered near object boundaries, whereas pixels located far from boundaries exhibit higher semantic stability. Therefore, for a specific queried class $c$, a changed pixel is considered reliable only if it maintains a safe distance from the edges in both temporal phases. 
% % changed
% We observate that boundary-induced candidates are usually concentrated near semantic boundaries, whereas reliable changed regions tend to contain pixels located inside the corresponding semantic regions. Therefore, for a queried class $c$, we estimate the spatial support of a candidate pixel by measuring its distance to the semantic boundaries in both temporal phases.}
To quantify this spatial support, we extract the undirected spatial boundaries of the binary masks $m_{1}^c$ and $m_{2}^c$, denoted as $B_{1}^c$ and $B_{2}^c$, respectively. We then compute the Euclidean Distance Transform (DT) from each pixel to the nearest boundary of its corresponding phase, yielding the distance maps $d_1^c, d_2^c \in \mathbb{R}^{H \times W}$.
When either $B_{1}^c$ or $B_{2}^c$  is empty, we set the corresponding distance map to $+\infty$, so that the distance from the available temporal observation is used.
We define the continuous spatial confidence $R^c(x)$ of a changing pixel as the minimum of its distances to the boundaries of both phases:
\begin{equation}
R^c(x) = \min \big( d_{1}^c(x), d_{2}^c(x) \big),
\end{equation}
where $R^c(x) \ge 0$ denotes the continuous distance field. Using the sigmoid function, we map this distance field to a smooth spatial confidence weight:
% \begin{equation}
% W^c(x) = \frac{1}{1 + \exp\left(-\frac{R^c(x) - \tau_c}{s}\right)},
% \end{equation}
\begin{equation}
W^c(x) = \frac{1}{1 + \exp\left(-(R^c(x) - \tau_c)\right)},
\end{equation}
where $W^c \in [0, 1]^{H \times W}$ defines the confidence weight map, and $\tau_c$ is the boundary tolerance threshold set for class $c$. 
% changed
When $W^c(x) > 0.5$, the pixel is treated as an interior-supported candidate pixel, indicating that it is sufficiently separated from the class-region boundaries in both temporal predictions.

% changed
While $W^c(x)$ provides a pixel-level spatial support measure, directly thresholding it may shrink candidate regions and remove valid boundary pixels. Therefore, we make the decision at the connected-component level. For each connected component $K_i$ within the category candidate region $M_{base}^c$, we calculate the proportion of interior-supported pixels. If this proportion exceeds a preset threshold $\rho \in (0, 1]$, the component is retained; otherwise, it is regarded as a boundary-dominated unreliable candidate and removed:

% origin
% While $W^c(x)$ provides a robust pixel-level confidence measure, applying a hard threshold directly would cause the boundaries of genuine change regions to contract, thereby compromising their structural integrity. To resolve this, we shift our perspective from isolated pixels to spatial regions. We assume that genuine changes must form spatially consistent regions with sufficient internal evidence, rather than existing solely as boundary artifacts. Therefore, we implement component-level decision-making. For any connected component $K_i$ within the category candidate region $M_{base}^c$, if the proportion of high-confidence core change pixels contained within $K_i$ exceeds a pre-set internal proportion threshold $\rho \in (0, 1]$, the entire connected component is deemed to represent a genuine structural change:
\begin{equation}
M_{out}^c(K_i) =
\begin{cases}
1, & \text{if } \frac{\sum_{x \in K_i} \mathbb{I}(W^c(x) > 0.5)}{|K_i|} \ge \rho \\
0, & \text{otherwise}
\end{cases},
\end{equation}
% changed
where $M_{out}^c \in \{0, 1\}^{H \times W}$ denotes the spatially filtered change mask for class $c$, and $|K_i|$ denotes the area (i.e., the number of pixels) of the connected component. This component-level criterion removes candidates dominated by boundary-induced responses while retaining regions that contain sufficient interior support. 
% Finally, the results for all categories are merged, followed by morphological opening and small-region removal to eliminate isolated outliers from the final binary mask.
Finally, morphological opening and small-component removal are applied to the resulting change mask to suppress isolated noise.

% origin
% where $M_{out}^c \in \{0, 1\}^{H \times W}$ forms the refined change mask for class $c$, and $|K_i|$ denotes the area (i.e., the number of pixels) of the connected component. This mechanism successfully suppresses false positives caused by edge misalignment while ensuring the complete topological structure of regions with genuine changes. Finally, as a post-processing step, the refinement results for all categories are merged, followed by morphological opening and small region removal to eliminate isolated outliers from the final binary mask. 

Since BCR is applied only to sparse candidate regions and uses lightweight distance-transform and connected-component operations, it introduces limited overhead compared with the backbone semantic parsing stage.

\begin{algorithm}[!h]
\caption{ReA-OVCD}
\label{alg:rea_ovcd}
\begin{algorithmic}[1]
\REQUIRE Bi-temporal images $I_1, I_2$; Target vocabulary $\mathcal{C}$; Parameters $\theta_{change}, \tau_c, \rho$.
\ENSURE Final per-class change masks $\{M_{out}^c\}_{c=1}^N$.

\STATE Use SAM-3 to extract pixel-wise semantic logits $\mathbf{L}_1, \mathbf{L}_2 \in \mathbb{R}^{N \times H \times W}$ and obtain discrete labels $Y_1, Y_2$;
\STATE Generate hard change mask: $H_{change} = \mathbb{I}(Y_1 \neq Y_2)$;
\STATE Compute JS divergence $D_{JS}$ and normalized logit difference $\tilde{D}_{logit}$ from $\mathbf{L}_1, \mathbf{L}_2$;
\STATE Compute fusion score: $S_{fusion} = \frac{1}{2} D_{JS} + \frac{1}{2}\,\tilde{D}_{logit}$;
\STATE Derive change score: $S_{change} = H_{change} \cdot S_{fusion}$;
\STATE Obtain base mask: $M_{base} = \mathbb{I}(S_{change} > \theta_{change})$;
\FOR{each class $c \in \{1, 2, \dots, N\}$}

    \STATE Compute class masks $m_{1}^c, m_{2}^c$ from $Y_1, Y_2$;

    \STATE Extract candidate mask:
    $M_{base}^c = M_{base} \cdot \mathbb{I}(m_{1}^c \neq m_{2}^c)$;

    \STATE Compute boundary distance maps $d_1^c, d_2^c$ via DT;

    \STATE Compute spatial confidence:
    $R^c = \min(d_1^c, d_2^c)$;

    \STATE Convert to confidence weights:
    $W^c = \sigma(R^c - \tau_c\big)$;

    \STATE Filter connected components $K_i \in M_{base}^c$:
    $M_{out}^c = \bigcup \left\{ K_i \;\middle|\;
    \frac{1}{|K_i|} \sum_{x \in K_i} \mathbb{I}(W^c(x) > 0.5) \ge \rho \right\}$;

\ENDFOR

\STATE \textbf{return} Final semantic change masks $\{M_{out}^c\}_{c=1}^N$
\end{algorithmic}
\end{algorithm}

% origin
% \subsection{Summary}
% Building upon the open-vocabulary semantic parsing capabilities of foundation models (e.g., SAM-3), ReA-OVCD adopts a pixel-level semantic modeling framework and introduces a collaborative refinement strategy to explicitly model change reliability.
% Specifically, starting from an initial change estimation derived from semantic label inconsistencies, candidate changes are first reassessed in the semantic domain by modeling distribution-level shifts and response variations, which reduces sensitivity to ambiguous predictions. This is followed by boundary-aware refinement in the spatial domain, where change regions are validated based on boundary-aware reliability and interior pixel support, thereby suppressing artifacts caused by misalignment and segmentation uncertainty.
% By integrating semantic reassessment and spatial validation into a unified pipeline, the proposed method goes beyond detecting simple appearance or disappearance events. It enables accurate identification of fine-grained local changes, such as partial building extensions, while mitigating the instability of naive pixel-wise comparisons.
% Ultimately, ReA-OVCD preserves the flexibility of pixel-level modeling while significantly improving the reliability and spatial consistency of change detection through a lightweight refinement strategy. The complete workflow is summarized in Algorithm~\ref{alg:rea_ovcd}.

% changed

\subsection{Summary}
ReA-OVCD revisits dense pixel-level OVCD by using label discrepancies as initial change indications and assessing their reliability from semantic and spatial perspectives. Instead of treating discrete label discrepancies as final change predictions, it uses them only to produce candidate change regions and then filters unreliable candidates with semantic and spatial evidence. SCR examines whether a candidate is supported by distribution-level divergence and response-level variation, reducing sensitivity to ambiguous open-vocabulary predictions. BCR further evaluates whether a candidate region contains sufficient interior support, removing boundary-dominated responses caused by segmentation uncertainty or slight geometric misalignment.
Through this design, ReA-OVCD preserves the flexible localization ability of pixel-level comparison while improving the reliability and spatial consistency of the resulting change masks. In addition, semantic parsing and SCR are shared across all queried categories, while only BCR is performed separately for each class. This avoids repeatedly executing the full OVCD pipeline for every category and unifies BCD and SCD, reducing the computational cost of multi-class semantic change detection. Since the proposed assessments are applied to candidate regions using lightweight tensor, distance-transform, and connected-component operations, the framework remains training-free and computationally efficient. The complete workflow is summarized in Algorithm~\ref{alg:rea_ovcd}.

\section{Dataset and Experimental Setup}
\subsection{Datasets}
\label{sec:datasets}
\textbf{LEVIR-CD}\cite{levircd} is a large-scale, high-resolution, dual-temporal remote sensing benchmark dataset for building change detection. The dataset is primarily derived from high-resolution commercial satellite imagery sourced from the Google Earth platform, with a spatial resolution of 0.5 meters. The imagery covers multiple urban areas in Texas, USA, between 2002 and 2018. The scenes feature significant seasonal variations in lighting and a diverse range of building types (such as large warehouses and dense residential estates), with a primary focus on the construction, extension and demolition of buildings. The dataset is divided into 445 training pairs, 64 validation pairs and 128 test pairs, with the image resolution maintained at $1024\times1024$ pixels.

\textbf{WHUCD}\cite{whucd} dataset is released by Wuhan University and is derived from an airborne high-resolution remote sensing platform. This dataset possesses extremely high spatial resolution (0.2 meters), providing exceptionally rich textural detail and geometric features. The scenes authentically document the surface topography and the damage to and reconstruction of buildings in the Christchurch region of New Zealand before and after the 2011 earthquake. The dataset focuses primarily on changes in large, relatively sparsely distributed building structures, with ideal lighting conditions and relatively simple, distinct variations. Following cropping and pre-processing, this paper constructed a test set comprising 690 pairs of images with a resolution of $512\times512$ pixels.

\textbf{DSIFN}\cite{dsifn} dataset is also constructed using high-resolution satellite imagery from the Google Earth platform, with a spatial resolution of approximately 2 meters. This dataset is highly representative of typical urban scenes, covering six major and medium-sized cities in China including Beijing, Shanghai and Shenzhen from 2000 to 2019. Compared to other datasets, the scene characteristics of DSIFN place greater emphasis on high-density urban expansion, featuring irregular shape variations, blurred boundaries, and significant interference noise (such as sensor noise), which places higher demands on the model’s robustness. The dataset is divided into 3,600 training pairs and 48 test pairs, all maintained at a resolution of $512\times512$.

\textbf{SECOND}\cite{second} is a remote sensing image dataset specifically designed for multi-class semantic change detection, with data sourced from multi-source high-resolution aerial and satellite remote sensing platforms. It features a wide range of spatial resolutions, spanning from 0.5 to 3 meters. The dataset covers multiple cities in China and provides detailed annotations of complex semantic transitions between six major land cover classes. Furthermore, the scenes incorporate real-world challenges such as boundary ambiguity and luminance noise. For the experiments, the image resolution was uniformly set to $512\times512$ pixels, and the dataset was split into 2,968 training pairs and 1,694 test pairs.

This paper selects a combination of test sets from the aforementioned four datasets to construct a comprehensive evaluation benchmark. These datasets not only span different sensor types (commercial satellites and aerial imagery) and spatial resolutions (0.2 m to 3 m), but their scene characteristics also span a wide spectrum, from the simple addition or removal of individual buildings, through dense urban expansion, to complex semantic transformations of multi-class land cover. By conducting validation directly on these test sets, we can evaluate the robustness of the proposed method in handling complex and variable remote sensing scenarios, as well as its generalization capability in performing open-vocabulary recognition and reasoning across diverse real-world contexts.

\begin{table}[htbp]
\centering
\caption{Text Prompt Construction for All Datasets. Each semantic class is expanded into multiple synonyms, each of which serves as an independent query.}
\label{tab:prompts}
\begin{tabular}{@{} l l p{5cm} @{}}
\toprule
\textbf{Dataset} & \textbf{Class} & \textbf{Text Prompts} \\
\midrule
\multirow{2}{*}{LEVIR-CD} 
& Background & ground, road, vegetation \\
& Building   & building, roof, house \\
\midrule
\multirow{2}{*}{WHU-CD} 
& Background & ground, road, vegetation \\
& Building   & building, roof, house \\
\midrule
\multirow{2}{*}{DSIFN} 
& Background & ground, road, vegetation, farmland \\
& Building   & building, house, roof, urban structure \\
\midrule
\multirow{7}{*}{SECOND} 
& Background & background \\
& Water      & water, river, lake, pond, reservoir, stream \\
& N.v.g. Surface & bare ground, bare soil, barren, dirt, sand \\
& Low Veg.   & low vegetation, grass, lawn, shrub, grassland, meadow \\
& Tree       & tree, trees, forest, woodland, canopy, grove \\
& Building   & building, roof, house, structure \\
& Playground & sports field, running track, athletic track, stadium \\
\bottomrule
\end{tabular}
\end{table}

\subsection{Text Prompt Construction}
Table~\ref{tab:prompts} outlines the text prompts constructed for all datasets. To enrich the semantic representation and capture the diverse visual appearances of geographical entities, we employ a lightweight synonym expansion strategy. For each semantic class (Column 2), we select common descriptors (Column 3) based on dataset priors, and each term serves as an independent query to interact with visual features.
Notably, this prompt construction is carefully tailored to dataset characteristics. To establish a clear semantic separation for the BCD datasets (LEVIR-CD, WHU-CD, DSIFN), we define background prompts using representative non-building land-cover categories (terms like \textit{ground}, \textit{road}, and \textit{vegetation}), allowing minor dataset-specific adaptations (e.g., adding \textit{farmland} for DSIFN). In contrast, the SECOND dataset inherently involves multi-class competition. Therefore, we expand foreground categories but strictly retain the raw \textit{background} label. This design effectively prevents semantic overlap and maintains mutual exclusivity among categories.

\subsection{Evaluation Metrics}
This paper adopts standard metrics from the change detection field and calculates them uniformly across change classes.  Based on the prediction results, we identify True Positives (TP) as correctly detected change pixels, False Positives (FP) as unchanged pixels incorrectly predicted as changes, and False Negatives (FN) as actual change pixels missed by the model. 

Accordingly, the Precision and Recall for the change class are defined as:
\begin{equation}
\text{Precision}^{C} = \frac{\text{TP}}{\text{TP} + \text{FP}}, \\
\text{Recall}^{C} = \frac{\text{TP}}{\text{TP} + \text{FN}}.
\end{equation}

To provide a more comprehensive assessment, we adopt $\mathrm{F}_1^{C}$ as the primary evaluation metric and $\text{IoU}^{C}$ as a key supplementary metric:
\begin{equation}
\mathrm{F}_1^{C} = \frac{2 \cdot \text{Precision}^{C} \cdot \text{Recall}^{C}}{\text{Precision}^{C} + \text{Recall}^{C}}, \\
\end{equation}

\begin{equation}
\text{IoU}^{C} = \frac{\text{TP}}{\text{TP} + \text{FP} + \text{FN}}.
\end{equation}
Here, $\mathrm{F}_1^{C}$ balances both precision and recall, providing a comprehensive reflection of detection completeness and false positive suppression capability. $\text{IoU}^{C}$ is more sensitive to region overlap and boundary alignment, enabling a more rigorous assessment of the spatial quality of the detected regions. For the SCD task, this paper calculates per-class $\mathrm{F}_1$ and per-class IoU for each change category, supplemented by category means as reference statistics. Considering the severe class imbalance inherent in change detection tasks, we adopt the $\mathrm{F}_1^{C}$ score as the primary evaluation metric, with $\text{IoU}^{C}$ serving as a supplementary standard metric.

\subsection{Experimental Design}
\subsubsection{Compared Methods}
% We conduct comparative experiments using a variety of representative algorithms, covering traditional zero-shot methods and state-of-the-art OVCD methods. Specifically, among the mainstream OVCD methods, we select open source AdaptOVCD as the primary baseline; simultaneously, two representative pipelines proposed in DynamicEarth are introduced: IMC (based on APE\cite{ape} and DINOv2\cite{dinov2}) and MCI (based on SAM, DINOv2 and SegEarth-OV\cite{segearthov}), as well as the latest OmniOVCD, UniVCD ,OpenDPR and Seg2Change models. Furthermore, the zero-shot comparison includes SCM, AnyChange, and BiSAM-CD to validate the effectiveness of our method under a training-free setting.
We conduct comparative experiments using a variety of representative algorithms, covering both state-of-the-art (SOTA) OVCD methods and traditional zero-shot approaches.
For the SOTA OVCD comparison, we select the open-source AdaptOVCD as our primary baseline. In addition, we include several other advanced models: two representative pipelines from DynamicEarth, namely IMC (based on APE~\cite{ape} and DINOv2~\cite{dinov2}) and MCI (based on SAM, DINOv2, and SegEarth-OV~\cite{segearthov}); and the latest models, including OmniOVCD, UniVCD, OpenDPR, and Seg2Change.
To validate the effectiveness of our method under a training-free setting, we further compare it against three zero-shot baselines: SCM, AnyChange, and BiSAM-CD.

\subsubsection{Quantitative Evaluation}
We conduct comprehensive quantitative evaluations on three widely used BCD benchmarks (DSIFN, LEVIR-CD, and WHU-CD) as well as a representative SCD dataset, SECOND. 
Since most existing zero-shot methods focus on BCD tasks, no comparison is made with them on the SCD task.

\subsubsection{Qualitative Evaluation}
To provide an intuitive comparison, we conduct a qualitative analysis against several representative  approaches that share the same task and have publicly available implementations. Specifically, we select typical samples from the BCD datasets and category-specific samples from the SECOND dataset to evaluate the semantic change recognition capabilities.

% \subsubsection{Efficiency Evaluation}
% We evaluate the computational efficiency of the proposed framework by comparing it with other training-free approaches under a unified hardware and experimental setting. We report latency and peak GPU memory consumption as efficiency metrics, together with $\mathrm{F}_{1}^{C}$ to reflect the relationship between detection performance and computational cost.}
% changed
\subsubsection{Efficiency Evaluation}
We evaluate the computational efficiency of the proposed framework by comparing it with other training-free approaches under a unified hardware and experimental setting. We report latency and peak GPU memory consumption as efficiency metrics.

\begin{table*}[!t]
\centering
\caption{Comparison with state-of-the-art methods on LEVIR-CD, WHU-CD, and DSIFN datasets. 
The \textcolor{best_red}{best} and \textcolor{second_blue}{second best} performances are highlighted. 
\colorbox{f1bg}{\strut $\mathrm{F}_{1}^{C}$} serves as the primary evaluation metric.
Precision (\colorbox{metricbg}{\strut Prec.}), Recall (\colorbox{metricbg}{\strut Rec.}), and \colorbox{metricbg}{\strut $\mathrm{IoU}^{C}$} are standard evaluation metrics. 
Methods marked with $^\dagger$ indicate results cited directly from their original papers.}
\label{tab:comparison_full}
\small
\renewcommand{\arraystretch}{1.2}
\setlength{\tabcolsep}{2.5pt}

\begin{tabularx}{\textwidth}{l|l|CCCC|CCCC|CCCC}
\toprule
\textbf{Method} & \textbf{Venue} 
& \multicolumn{4}{c|}{\textbf{LEVIR-CD (\%)}} 
& \multicolumn{4}{c|}{\textbf{WHU-CD (\%)}} 
& \multicolumn{4}{c}{\textbf{DSIFN (\%)}} \\

& & \cellcolor{metricbg}Prec. 
& \cellcolor{metricbg}Rec. 
& \cellcolor{metricbg}$\mathrm{IoU}^{C}$ 
& \cellcolor{f1bg}$\mathrm{F}_{1}^{C}$ 

& \cellcolor{metricbg}Prec. 
& \cellcolor{metricbg}Rec. 
& \cellcolor{metricbg}$\mathrm{IoU}^{C}$ 
& \cellcolor{f1bg}$\mathrm{F}_{1}^{C}$ 

& \cellcolor{metricbg}Prec. 
& \cellcolor{metricbg}Rec. 
& \cellcolor{metricbg}$\mathrm{IoU}^{C}$ 
& \cellcolor{f1bg}$\mathrm{F}_{1}^{C}$ \\
\midrule

\multicolumn{13}{l}{\textit{Traditional zero-shot methods}} \\
Anychange\cite{anychange} & NeurIPS 24 & 20.46 & 81.06 & 19.53 & 32.68 & 13.67 & \textbf{\textcolor{best_red}{95.43}} & 13.58 & 23.91 & 32.80 & 50.89 & 24.91 & 39.89 \\
SCM\cite{scm} & IGARSS 24 & 23.99 & 48.73 & 19.15 & 32.15 & 24.55 & 63.98 & 21.57 & 35.49 & 42.43 & 20.27 & 15.89 & 27.43 \\
BiSAM-CD(GT)\cite{bisamcd} & TGRS 25 & \textbf{\textcolor{best_red}{78.89}} & 57.87 & 50.11 & 66.76 & \textbf{\textcolor{best_red}{93.07}} & 79.46 & \textbf{\textcolor{second_blue}{75.02}} & \textbf{\textcolor{second_blue}{85.72}} & \textbf{\textcolor{best_red}{93.55}} & 12.23 & 12.13 & 21.63 \\
\midrule

\multicolumn{13}{l}{\textit{Open-vocabulary methods}} \\
IMC\cite{dynamicearth} & AAAI 26 & 58.45 & 65.32 & 44.61 & 61.70 & 87.26 & 76.09 & 68.48 & 81.29 & \textbf{\textcolor{second_blue}{90.97}} & 15.46 & 15.22 & 26.42 \\
MCI\cite{dynamicearth} & AAAI 26 & 45.75 & 64.05 & 36.40 & 53.38 & 63.41 & 67.93 & 48.80 & 65.59 & 68.87 & 44.22 & 36.85 & 53.86 \\
% UniVCD$^\dagger$\cite{univcd} & arXiv 25 & 65.10 & \textbf{\textcolor{second_blue}{81.40}} & 56.70 & 72.30 & 73.30 & 86.80 & 66.00 & 79.50 & -- & -- & -- & -- \\
%changed
UniVCD$^\dagger$\cite{univcd} & arXiv 25 & 65.10 & 81.40 & 56.70 & 72.30 & 73.30 & 86.80 & 66.00 & 79.50 & -- & -- & -- & -- \\
OmniOVCD$^\dagger$\cite{omniovcd} & arXiv 26 & -- & -- & \textbf{\textcolor{second_blue}{67.20}} & \textbf{\textcolor{second_blue}{80.40}} & -- & -- & 66.50 & 79.90 & -- & -- & -- & -- \\
OpenDPR$^\dagger$\cite{opendpr} & CVPR 26 & -- & -- & 44.80 & 61.90 & -- & -- & 54.30 & 70.40 & -- & -- & -- & -- \\
% AdaptOVCD\cite{adaptovcd} & arXiv 26 & 62.83 & 74.10 & 51.52 & 68.00 & 80.60 & 72.85 & 61.99 & 76.53 & 53.34 & \textbf{\textcolor{second_blue}{63.64}} & \textbf{\textcolor{second_blue}{40.88}} & \textbf{\textcolor{second_blue}{58.04}} \\
%changed
AdaptOVCD\cite{adaptovcd} & arXiv 26 & 62.83 & 74.10 & 51.52 & 68.00 & 80.60 & 72.85 & 61.99 & 76.53 & 53.34 & \textbf{\textcolor{second_blue}{63.64}} & 40.88 & 58.04 \\
%added
Seg2Change\cite{seg2change} & arXiv 26 & 69.70 & \textbf{\textcolor{second_blue}{90.40}} & 64.89 & 78.71 & 80.15 & 92.00 & 74.93 & 85.67 & 62.63 & 56.37 & \textbf{\textcolor{second_blue}{42.18}} & \textbf{\textcolor{second_blue}{59.34}} \\
\midrule

% \textbf{ours} & -- & \textbf{\textcolor{second_blue}{74.00}} & \textbf{\textcolor{best_red}{93.28}} & \textbf{\textcolor{best_red}{70.26}} & \textbf{\textcolor{best_red}{82.53}} & \textbf{\textcolor{second_blue}{88.48}} & \textbf{\textcolor{second_blue}{92.21}} & \textbf{\textcolor{best_red}{82.33}} & \textbf{\textcolor{best_red}{90.31}} & 62.28 & \textbf{\textcolor{best_red}{74.36}} & \textbf{\textcolor{best_red}{51.27}} & \textbf{\textcolor{best_red}{67.79}} \\
%changed
\textbf{ours} & -- & \textbf{\textcolor{second_blue}{77.74}} & \textbf{\textcolor{best_red}{92.60}} & \textbf{\textcolor{best_red}{73.19}} & \textbf{\textcolor{best_red}{84.52}} & \textbf{\textcolor{second_blue}{88.48}} & \textbf{\textcolor{second_blue}{92.21}} & \textbf{\textcolor{best_red}{82.33}} & \textbf{\textcolor{best_red}{90.31}} & 62.28 & \textbf{\textcolor{best_red}{74.36}} & \textbf{\textcolor{best_red}{51.27}} & \textbf{\textcolor{best_red}{67.79}} \\
\bottomrule
\end{tabularx}
\end{table*}

\subsubsection{Ablation Study}
To validate the effectiveness of the proposed framework, we conduct a progressive ablation study from the overall architecture to the internal module designs. We further analyze intermediate representations via qualitative visualizations to better understand the role of each module.

\subsection{Experimental Settings}
All experiments were conducted on a single NVIDIA RTX 3090 GPU, with the algorithms implemented using the PyTorch framework. This paper utilizes the official open-sourced SAM-3 model, which by default resizes all input images to a resolution of 1008×1008. 
% origin
% In terms of input resolution, both our method and the IMC pipeline employ a $512\times512$ sliding window for inference on the LEVIR-CD dataset; all other experiments follow Section \ref{sec:datasets}.
% changed
Due to input-size constraints, our method, IMC, and Seg2Change process each pair with four non-overlapping 512$\times$512 windows and stitch the predictions for evaluation on the LEVIR-CD dataset. All other experiments follow the resolutions in Section~\ref{sec:datasets}.
To ensure a fair comparison, each baseline retains its original configuration and core algorithmic workflow, with only necessary compatibility adjustments made. 
For our method, the hyperparameter $\rho$ is fixed across all datasets. Other parameters are kept consistent whenever possible, with only minimal dataset-specific adjustments to threshold-related values when necessary to account for characteristics of each dataset.

\section{Experimental Results}

\subsection{Quantitative Evaluation}
To comprehensively evaluate the effectiveness of the proposed ReA-OVCD, we compare it against traditional zero-shot methods and recent OVCD approaches across multiple benchmark datasets. 

\subsubsection{Building Change Detection}
As shown in Table~\ref{tab:comparison_full}, we conduct a quantitative comparison on three binary building change detection datasets with distinct scene characteristics. ReA-OVCD achieves the best $\mathrm{F}_{1}^{C}$ and $\mathrm{IoU}^{C}$ on all three datasets, demonstrating the effectiveness of semantic-spatial reliability assessment under diverse building-change scenarios.
On WHU-CD, where buildings are relatively large and sparsely distributed, ReA-OVCD achieves an $\mathrm{F}_{1}^{C}$ of $90.31\%$ and an $\mathrm{IoU}^{C}$ of $82.33\%$, not only outperforming Seg2Change by $4.64\%$ in $\mathrm{F}_{1}^{C}$ but also surpassing the traditional zero-shot method BiSAM-CD that relies on ground-truth prompts.
This result benefits from flexible candidate generation and BCR's removal of boundary-dominated responses in sparse scenes.
For LEVIR-CD, which contains denser building changes, ReA-OVCD obtains the highest $\mathrm{F}_{1}^{C}$ of $84.52\%$ and $\mathrm{IoU}^{C}$ of $73.19\%$, exceeding OmniOVCD by $4.12\%$ in $\mathrm{F}_{1}^{C}$. The improvement indicates that assessing reliability can retain localized building changes while filtering unreliable responses around adjacent structures.
On DSIFN, where low resolution, blurred boundaries, and environmental noise make semantic comparison less stable, ReA-OVCD maintains an $\mathrm{F}_{1}^{C}$ of $67.79\%$, improving over Seg2Change by $8.45\%$. This robustness comes from using label discrepancies only as candidate changes and then filtering unreliable candidates with semantic response evidence and spatial interior support, rather than directly trusting discretized pixel-wise predictions.

\begin{table*}[!t]
\centering
\caption{Per-class comparison on the SECOND dataset. 
The \textcolor{best_red}{best} and \textcolor{second_blue}{second best} performances are highlighted.
\colorbox{f1bg}{\strut $\mathrm{F}_{1}^{C}$} serves as the primary evaluation metric and \colorbox{metricbg}{\strut $\mathrm{IoU}^{C}$} is a standard evaluation metric. 
Methods marked with $^\dagger$ indicate results cited directly from their original papers.}
\label{tab:second_per_class}
\small
\renewcommand{\arraystretch}{1.2}
\setlength{\tabcolsep}{2.5pt}

\begin{tabularx}{\textwidth}{l|CC|CC|CC|CC|CC|CC|CC}
\toprule
\textbf{Method} 
& \multicolumn{2}{c|}{\textbf{Building}} 
& \multicolumn{2}{c|}{\textbf{Tree}} 
& \multicolumn{2}{c|}{\textbf{Water}} 
& \multicolumn{2}{c|}{\textbf{Low veg.}} 
& \multicolumn{2}{c|}{\textbf{N.v.g. Surface}} 
& \multicolumn{2}{c|}{\textbf{Playground}} 
& \multicolumn{2}{c}{\textbf{Avg.}} \\

& \cellcolor{metricbg}$\mathrm{IoU}^{C}$ & \cellcolor{f1bg}$\mathrm{F}_{1}^{C}$ 
& \cellcolor{metricbg}$\mathrm{IoU}^{C}$ & \cellcolor{f1bg}$\mathrm{F}_{1}^{C}$ 
& \cellcolor{metricbg}$\mathrm{IoU}^{C}$ & \cellcolor{f1bg}$\mathrm{F}_{1}^{C}$ 
& \cellcolor{metricbg}$\mathrm{IoU}^{C}$ & \cellcolor{f1bg}$\mathrm{F}_{1}^{C}$ 
& \cellcolor{metricbg}$\mathrm{IoU}^{C}$ & \cellcolor{f1bg}$\mathrm{F}_{1}^{C}$ 
& \cellcolor{metricbg}$\mathrm{IoU}^{C}$ & \cellcolor{f1bg}$\mathrm{F}_{1}^{C}$ 
& \cellcolor{metricbg}$\mathrm{IoU}^{C}$ & \cellcolor{f1bg}$\mathrm{F}_{1}^{C}$ \\
\midrule

IMC & 31.74 & 48.19 & 10.89 & 19.64 & 12.34 & 21.97 & 0.40 & 0.79 & 0.00 & 0.00 & 26.93 & 42.43 & 13.72 & 22.17 \\
MCI & 38.65 & 55.75 & 15.38 & 26.66 & 15.39 & 26.68 & 21.17 & 34.95 & 27.91 & 43.65 & 22.39 & 36.59 & 23.48 & 37.38 \\
% UniVCD$^\dagger$ & 43.20 & 60.40 & 18.90 & 31.90 & 8.20 & 15.20 & \textbf{\textcolor{second_blue}{24.90}} & \textbf{\textcolor{second_blue}{39.80}} & 28.00 & 43.70 & 0.00 & 0.00 & 20.53 & 31.83 \\
%changed
UniVCD$^\dagger$ & 43.20 & 60.40 & 18.90 & 31.90 & 8.20 & 15.20 & \textbf{\textcolor{second_blue}{24.90}} & \textbf{\textcolor{second_blue}{39.80}} & 28.00 & 43.70 & 0.00 & 0.00 & 20.53 & 31.83 \\
% OmniOVCD$^\dagger$ & \textbf{\textcolor{second_blue}{45.20}} & \textbf{\textcolor{second_blue}{62.30}} & 16.70 & 28.60 & 21.20 & 35.00 & 24.50 & 39.30 & 27.70 & 43.40 & 27.00 & 42.40 & 27.05 & 41.83 \\
%changed
OmniOVCD$^\dagger$ & 45.20 & 62.30 & 16.70 & 28.60 & 21.20 & 35.00 & 24.50 & 39.30 & 27.70 & 43.40 & 27.00 & 42.40 & 27.05 & 41.83 \\
OpenDPR$^\dagger$ & 42.40 & 59.50 & \textbf{\textcolor{best_red}{20.90}} & \textbf{\textcolor{best_red}{34.50}} & 17.50 & 29.80 & 23.00 & 37.40 & 30.20 & 46.40 & \textbf{\textcolor{best_red}{37.30}} & \textbf{\textcolor{best_red}{54.30}} & 28.55 & \textbf{\textcolor{second_blue}{43.65}} \\
% AdaptOVCD & 44.98 & 62.05 & 12.67 & 22.49 & \textbf{\textcolor{second_blue}{23.34}} & \textbf{\textcolor{second_blue}{37.84}} & 21.21 & 34.99 & \textbf{\textcolor{second_blue}{33.99}} & \textbf{\textcolor{second_blue}{50.74}} & \textbf{\textcolor{second_blue}{29.28}} & \textbf{\textcolor{second_blue}{45.30}} & 27.58 & 42.24 \\
%changed
AdaptOVCD & 44.98 & 62.05 & 12.67 & 22.49 & \textbf{\textcolor{second_blue}{23.34}} & \textbf{\textcolor{second_blue}{37.84}} & 21.21 & 34.99 & 33.99 & 50.74 & \textbf{\textcolor{second_blue}{29.28}} & \textbf{\textcolor{second_blue}{45.30}} & 27.58 & 42.24 \\
%added
Seg2Change & \textbf{\textcolor{best_red}{61.77}} & \textbf{\textcolor{best_red}{76.37}} & 16.05 & 27.66 & 23.04 & 37.46 & 18.16 & 30.74 & \textbf{\textcolor{best_red}{37.26}} & \textbf{\textcolor{best_red}{54.29}} & 17.73 & 30.13 & \textbf{\textcolor{second_blue}{29.00}} & 42.78 \\
\midrule

% \textbf{ours} & \textbf{\textcolor{best_red}{59.12}} & \textbf{\textcolor{best_red}{74.31}} & \textbf{\textcolor{second_blue}{19.40}} & \textbf{\textcolor{second_blue}{32.48}} & \textbf{\textcolor{best_red}{26.09}} & \textbf{\textcolor{best_red}{41.39}} & \textbf{\textcolor{best_red}{27.95}} & \textbf{\textcolor{best_red}{43.69}} & \textbf{\textcolor{best_red}{34.80}} & \textbf{\textcolor{best_red}{51.63}} & 24.74 & 39.67 & \textbf{\textcolor{best_red}{32.02}} & \textbf{\textcolor{best_red}{47.19}} \\
%changed
\textbf{ours} & \textbf{\textcolor{second_blue}{59.12}} & \textbf{\textcolor{second_blue}{74.31}} & \textbf{\textcolor{second_blue}{19.40}} & \textbf{\textcolor{second_blue}{32.48}} & \textbf{\textcolor{best_red}{26.09}} & \textbf{\textcolor{best_red}{41.39}} & \textbf{\textcolor{best_red}{27.95}} & \textbf{\textcolor{best_red}{43.69}} & \textbf{\textcolor{second_blue}{34.80}} & \textbf{\textcolor{second_blue}{51.63}} & 24.74 & 39.67 & \textbf{\textcolor{best_red}{32.02}} & \textbf{\textcolor{best_red}{47.19}} \\

\bottomrule
\end{tabularx}
\end{table*}

\subsubsection{Land Cover Change Detection}
Unlike binary building detection, multi-class semantic change detection requires the model to locate changes and identify category-specific semantic transitions. Table~\ref{tab:second_per_class} reports the per-class results on SECOND. ReA-OVCD achieves the best average $\mathrm{F}_{1}^{C}$ and $\mathrm{IoU}^{C}$ of $47.19\%$ and $32.02\%$, improving average $\mathrm{F}_{1}^{C}$ by $3.54\%$ over OpenDPR.
At the class level, ReA-OVCD obtains competitive performance on structurally coherent categories such as \textit{Building}, reaching an $\mathrm{F}_{1}^{C}$ of $74.31\%$. For land-cover categories with strong spatial continuity and semantic ambiguity, ReA-OVCD achieves the best performance on \textit{Water} and \textit{Low vegetation}, and remains competitive on \textit{N.v.g. Surface}. These results suggest that semantic response evidence helps suppress ambiguous label flips, while region-level spatial support helps preserve coherent candidate regions.
% For \textit{Playground}, the main difficulty comes from the initial SAM-3 semantic parsing stage. In the full-category setting, newly built playgrounds or grass-covered sports fields are often assigned higher responses for \textit{Low vegetation} or \textit{N.v.g. Surface} than for \textit{Playground}, which weakens the target activation before SCR and BCR are applied. The low Playground score of Seg2Change, which is also built on SAM-3, further suggests that this category is challenging for the underlying parser. Methods with stronger Playground scores commonly evaluate each category in a binary prompt setting, such as \textit{background} versus \textit{Playground}, which reduces category competition but requires repeated forward passes for different classes. In contrast, ReA-OVCD shares semantic parsing and SCR across all categories and performs only BCR separately, so its main forward cost does not grow linearly with the number of classes.
For \textit{Playground}, the main difficulty comes from the initial SAM-3 semantic parsing stage. In the full-category setting, newly built playgrounds or grass-covered sports fields can be assigned higher responses for \textit{Low vegetation} or \textit{N.v.g. Surface}. Therefore, although the proposed reliability assessment improves the result over direct label comparison, it cannot fully recover a category whose initial activation is insufficient.
In contrast, several competing methods evaluate each category in a binary prompt setting, such as \textit{background} versus \textit{Playground}, which can strengthen the response of a specific target class by reducing category competition, but requires repeated inference for different categories.
This reflects a trade-off between class-wise target activation and shared full-category inference. ReA-OVCD preserves inter-class competition to obtain balanced multi-class results, while sharing semantic parsing and SCR across all queried categories and performing only BCR separately. As a result, it maintains better overall performance with lower category-dependent computational cost.

\subsection{Qualitative Evaluation}

\begin{figure*}[!t]
\centering
\includegraphics[width=\textwidth]{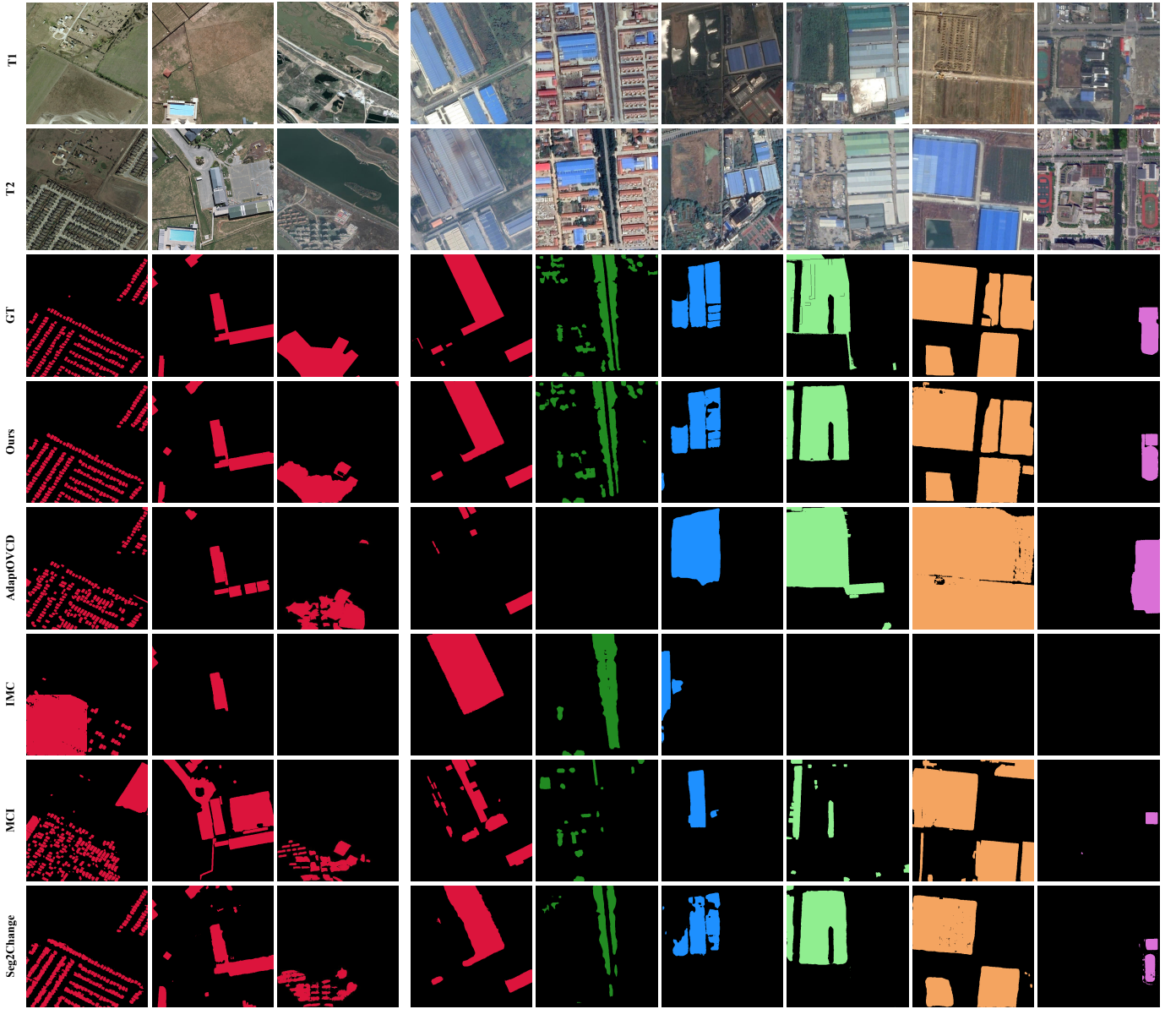}
\caption{Qualitative analysis on LEVIR-CD, WHU-CD, DSIFN, and SECOND (left part for the three BCD datasets, right part for SECOND). From top to bottom, each row shows: Image 1 (T1), Image 2 (T2), Ground Truth (GT), our method (Ours), AdaptOVCD, IMC, MCI and Seg2Change. Color legend: 
\textcolor[HTML]{DC143C}{\rule{0.6em}{0.6em}} Building,
\textcolor[HTML]{228B22}{\rule{0.6em}{0.6em}} Tree,
\textcolor[HTML]{1E90FF}{\rule{0.6em}{0.6em}} Water,
\textcolor[HTML]{90EE90}{\rule{0.6em}{0.6em}} Low Vegetation,
\textcolor[HTML]{F4A460}{\rule{0.6em}{0.6em}} N.v.g. Surface,
\textcolor[HTML]{DA70D6}{\rule{0.6em}{0.6em}} Playground.}
\label{fig:qualitative_all}
\end{figure*}
As shown in Figure~\ref{fig:qualitative_all}, we present a unified qualitative comparison across four datasets, including LEVIR-CD, WHU-CD, DSIFN, and SECOND. This visualization enables a direct comparison of different methods under consistent visual contexts, facilitating a comprehensive evaluation of change detection quality across diverse scenarios.

For the three BCD datasets, ReA-OVCD produces building change masks that are generally more spatially coherent while retaining localized changes.
On LEVIR-CD, AdaptOVCD misses several real changed regions, while IMC and MCI not only suffer from missed detections but also tend to merge adjacent buildings into large connected regions. Seg2Change provides visually comparable detection coverage, but nearby small buildings are still sometimes connected. In contrast, ReA-OVCD achieves robust building change identification with fewer missed regions and avoids the erroneous merging of adjacent buildings. This is mainly because pixel-wise candidate generation preserves localized building changes, while SCR filters unreliable semantic discrepancies that may otherwise connect nearby candidates.
On WHU-CD, ReA-OVCD detects not only complete building additions but also partial building extensions. This reflects the flexibility of pixel-level candidate generation, since several competing methods tend to miss such localized changes. Seg2Change can also capture most building changes, but still introduces several scattered false positives in unchanged areas; ReA-OVCD reduces these unreliable responses through semantic response evidence and spatial reliability validation.
For DSIFN, where buildings and vegetation are often interwoven and boundaries are less clear, ReA-OVCD more accurately recovers complete changed building regions while suppressing false responses in unchanged areas. This shows that the proposed candidate-validation process improves the robustness of dense pixel-level OVCD under low-resolution and noisy semantic predictions.

For the semantic change detection dataset SECOND (right part of Figure~\ref{fig:qualitative_all}), we further evaluate performance across six land-cover categories. Overall, ReA-OVCD exhibits stable visual performance across categories with diverse spatial characteristics. 
For the \textit{Building} category, ReA-OVCD identifies both regular expansions and more complex structural changes, while reducing scattered false positives compared with Seg2Change. 
For the \textit{Tree} category, ReA-OVCD captures both large contiguous regions and scattered small-scale changes, whereas some changed regions are missed by competing methods. 
For the \textit{Water} category, where changes often appear as gradual boundary shifts, ReA-OVCD produces more continuous changed regions, while Seg2Change may generate fragmented predictions. 
These results suggest that pixel-wise candidate generation provides the flexibility needed for irregular semantic transitions, while semantic-spatial reliability validation helps remove unstable dense responses and preserve coherent changed regions.
Similarly, for \textit{Low vegetation} and \textit{N.v.g. Surface}, ReA-OVCD better preserves local structural details and avoids turning ambiguous surrounding areas into small false positives, benefiting from semantic response evidence under multi-class category competition. 
For \textit{Playground}, although it remains challenging due to its visual similarity to grass-like or ground-like regions, ReA-OVCD produces cleaner regions with fewer fragmented false positives, indicating that reliability validation can reduce unreliable candidates even when the initial category activation is ambiguous.

Overall, the qualitative results support the main design motivation of ReA-OVCD. Pixel-wise candidate generation provides the flexibility needed to capture localized and irregular changes, while semantic-spatial reliability validation improves the reliability of these dense candidates. By using semantic response evidence and region-level spatial support, ReA-OVCD reduces unreliable label-discrepancy responses and boundary-induced artifacts while preserving fine-grained change localization in both BCD and SCD scenarios.

% changed
% Overall, the qualitative analysis intuitively demonstrates that the proposed semantic–spatial collaborative refinement strategy effectively improves change detection reliability by suppressing unstable semantic discrepancies and preserving spatially consistent change patterns.

\subsection{Efficiency Analysis}

We conduct a systematic efficiency comparison with representative change detection approaches under the same hardware setting using a single NVIDIA RTX 3090 GPU. To avoid conflating computational cost with differences in input resolution, all methods are evaluated using 512$\times$512 image pairs. We report the average end-to-end latency per image pair and peak GPU memory consumption, while detection accuracy is presented in the quantitative comparison tables above. The evaluation covers WHU-CD, LEVIR-CD and DSIFN, which contain building changes with different object densities, as well as the multi-class SECOND dataset. For SECOND, latency is the average per-pair time required to produce predictions for all queried categories. Peak GPU memory is measured as the maximum observed across sequential category-wise runs for the comparison methods and during joint inference for ReA-OVCD.

\begin{table*}[!t]
\centering
\caption{
Efficiency comparison on 512$\times$512 image pairs. Latency is reported in seconds per pair, and memory denotes peak GPU memory consumption in GB. Best results for each dataset are highlighted in \textbf{bold}.
}
\label{tab:efficiency_comparison}
\small
\renewcommand{\arraystretch}{1.15}
\setlength{\tabcolsep}{4.5pt}
\begin{tabular}{lcccccccc}
\toprule
\multirow{2}{*}{\textbf{Method}}
& \multicolumn{2}{c}{\textbf{WHU-CD}}
& \multicolumn{2}{c}{\textbf{LEVIR-CD}}
& \multicolumn{2}{c}{\textbf{DSIFN}}
& \multicolumn{2}{c}{\textbf{SECOND}} \\
\cmidrule(lr){2-3}
\cmidrule(lr){4-5}
\cmidrule(lr){6-7}
\cmidrule(lr){8-9}
& \makecell{\textbf{Latency}\\\textbf{(s)}}
& \makecell{\textbf{Memory}\\\textbf{(GB)}}
& \makecell{\textbf{Latency}\\\textbf{(s)}}
& \makecell{\textbf{Memory}\\\textbf{(GB)}}
& \makecell{\textbf{Latency}\\\textbf{(s)}}
& \makecell{\textbf{Memory}\\\textbf{(GB)}}
& \makecell{\textbf{Latency}\\\textbf{(s)}}
& \makecell{\textbf{Memory}\\\textbf{(GB)}} \\
\midrule
IMC
& 1.58 & 12.85
& 1.53 & 13.46
& 1.61 & 12.24
& 11.09 & 18.96 \\

MCI
& 2.10 & 6.55
& 4.80 & 6.55
& 5.34 & 6.55
& 27.17 & 8.77 \\

AdaptOVCD
& 6.14 & 10.57
& 16.85 & 10.57
& 24.35 & 10.57
& 73.83 & 15.17 \\

Seg2Change
& 1.75 & 6.03
& 1.75 & 6.08
& 2.17 & 6.05
& 17.46 & 9.11 \\

\textbf{Ours}
& \textbf{0.94} & \textbf{5.53}
& \textbf{0.95} & \textbf{5.56}
& \textbf{1.09} & \textbf{5.49}
& \textbf{4.30} & \textbf{8.34} \\
\bottomrule
\end{tabular}
\end{table*}

As shown in Table~\ref{tab:efficiency_comparison}, ReA-OVCD achieves the lowest latency and peak GPU memory consumption on all four datasets. On WHU-CD, ReA-OVCD processes each image pair in 0.94~s while using 5.53~GB of peak GPU memory. On the denser LEVIR-CD scenes, its latency increases only slightly to 0.95~s, with a peak memory consumption of 5.56~GB. This stable behavior indicates that the additional cost introduced by candidate filtering remains limited even when the candidate regions become denser and more complex.
The efficiency advantage is also evident on SECOND and DSIFN. On SECOND, where the latency includes the complete six-class inference process, ReA-OVCD requires 4.30~s per sample and 8.34~GB of peak GPU memory. In comparison, the fastest competing method, DynamicEarth with APE-DINOv2, requires 11.09~s per sample, corresponding to a $2.58\times$ speedup for ReA-OVCD. ReA-OVCD also consumes less memory than the most memory-efficient competing configuration, which requires 8.77~GB. On DSIFN, ReA-OVCD achieves a latency of 1.09~s and a peak memory consumption of 5.49~GB. Compared with the best competing results of 1.61~s and 6.05~GB in the respective metrics, ReA-OVCD provides both faster inference and lower memory consumption.

This computational efficiency mainly arises from the unified semantic parsing and lightweight semantic-spatial assessment design. The framework avoids repeated multi-model feature extraction and costly cross-instance association. Specifically, SCR is implemented using lightweight tensor operations, while BCR performs distance-transform and connected-component analysis only within the generated candidate regions. Moreover, in multi-class scenarios, semantic parsing and SCR are computed once for each image pair and shared across all queried categories, whereas only the lightweight BCR procedure is applied separately to the candidate regions of each category. This sharing mechanism limits category-dependent computational overhead compared with class-wise binary prompting protocols. Consequently, ReA-OVCD maintains favorable latency and memory efficiency alongside its improved detection performance, including when producing predictions for multiple semantic change categories on SECOND.

\subsection{Ablation Study}
To systematically validate the effectiveness of the proposed ReA-OVCD framework, we conduct progressive ablation studies at both the module level and the intra-module level. Specifically, we analyze the individual and joint contributions of the SCR and the BCR modules from the perspective of semantic and spatial reliability.

\begin{table}[!h]
\centering
\caption{Ablation study of integrating SCR and BCR. Results on SECOND are reported as class-averaged scores. Best performance on each dataset is shown in \textbf{bold}.}
\label{tab:overall_ablation}
\small
\renewcommand{\arraystretch}{1.2}
\setlength{\tabcolsep}{3pt}

\resizebox{\columnwidth}{!}{
\begin{tabular}{l|cc|cc|cc|cc}
\toprule
\textbf{Method} 
& \multicolumn{2}{c|}{\textbf{LEVIR-CD}} 
& \multicolumn{2}{c|}{\textbf{WHU-CD}} 
& \multicolumn{2}{c|}{\textbf{DSIFN}} 
& \multicolumn{2}{c}{\textbf{SECOND avg.}} \\

& \cellcolor{metricbg}$\mathrm{IoU}^{C}$ & \cellcolor{f1bg}$\mathrm{F}_{1}^{C}$ 
& \cellcolor{metricbg}$\mathrm{IoU}^{C}$ & \cellcolor{f1bg}$\mathrm{F}_{1}^{C}$ 
& \cellcolor{metricbg}$\mathrm{IoU}^{C}$ & \cellcolor{f1bg}$\mathrm{F}_{1}^{C}$ 
& \cellcolor{metricbg}$\mathrm{IoU}^{C}$ & \cellcolor{f1bg}$\mathrm{F}_{1}^{C}$ \\
\midrule

Baseline (XOR) 
% & 54.52 & 70.57 & 48.26 & 65.10 & 40.36 & 57.51 & 27.24 & 41.49 \\
%changed
& 53.28 & 69.52 & 48.26 & 65.10 & 40.36 & 57.51 & 27.24 & 41.49 \\

SCR only 
% & 61.67 & 76.29 & 62.99 & 77.29 & 49.01 & 65.78 & 29.64 & 44.52 \\
%changed
& 64.38 & 78.33 & 62.99 & 77.29 & 49.01 & 65.78 & 29.64 & 44.52 \\

BCR only 
% & 65.55 & 79.19 & 82.31 & 90.30 & 43.14 & 60.27 & 30.29 & 45.04 \\
%changed
& 70.79 & 82.90 & 82.31 & 90.30 & 43.14 & 60.27 & 30.29 & 45.04 \\

SCR + BCR 
% & \textbf{70.26} & \textbf{82.53} 
%changed
& \textbf{73.19} & \textbf{84.52} 
& \textbf{82.33} & \textbf{90.31} 
& \textbf{51.27} & \textbf{67.79} 
& \textbf{32.02} & \textbf{47.19} \\

\bottomrule
\end{tabular}
}
\end{table}

\subsubsection{Overall Module Integration}
As shown in Table~\ref{tab:overall_ablation}, we establish a baseline by directly applying a logical XOR operation on the bi-temporal discrete semantic labels. Introducing SCR alone improves $\mathrm{F}_{1}^{C}$ from $69.52\%$ to $78.3\%$ on LEVIR-CD and from $65.10\%$ to $77.29\%$ on WHU-CD, indicating that continuous semantic evidence helps filter unreliable label-discrepancy candidates caused by ambiguous predictions.
When BCR is applied independently, performance also improves on most datasets. The gains are especially clear on WHU-CD and LEVIR-CD, where candidate changes are strongly affected by boundary uncertainty and geometric misalignment. This suggests that evaluating region-level interior support helps remove boundary-dominated candidates while preserving spatially coherent regions.
Combining SCR and BCR leads to the best overall performance, reaching the highest scores across all datasets. These results show that semantic evidence and spatial support address complementary sources of unreliability in dense pixel-level OVCD.

\begin{figure}[!t]
\centering
\includegraphics[width=3.5in]{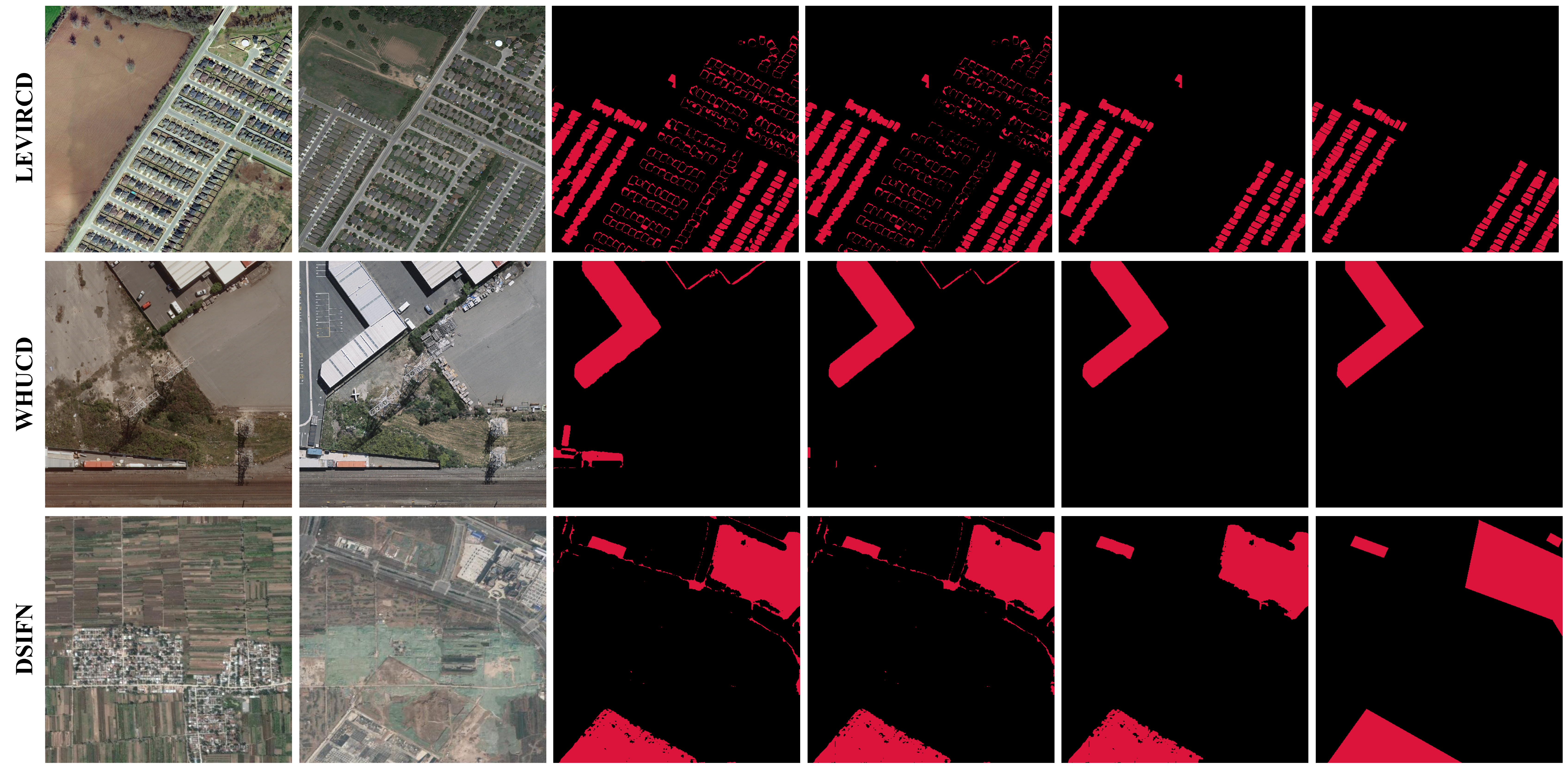}
\caption{Qualitative analysis of the proposed SCR and BCR modules through intermediate visualizations. 
From left to right, each row shows Image 1 (T1), Image 2 (T2), the initial hard change map, the result after applying SCR, the final prediction with both SCR and BCR (Ours), and the ground truth (GT). 
From top to bottom, examples are selected from LEVIR-CD, WHU-CD, and DSIFN, respectively. }
\label{fig:qualitative_3}
\end{figure}

To further understand the distinct roles of SCR and BCR, we visualize intermediate results in Figure~\ref{fig:qualitative_3}.
From the visual comparisons, we observe that the two modules address fundamentally different failure modes rooted in semantic and spatial unreliability.
First, SCR reduces false positives caused by unstable semantic predictions. 
As shown in the second row, the initial hard change map (Column 3) produces spurious responses in unchanged regions due to label fluctuations near decision boundaries. 
After applying SCR (Column 4), these responses are effectively suppressed, leading to more reliable semantic comparisons.
Second, BCR alleviates boundary artifacts in the change maps. 
As observed in the first and third rows, the initial predictions exhibit band-like responses along object boundaries, mainly caused by slight misregistration and segmentation uncertainty. 
After BCR is further integrated (Column 5), these boundary artifacts are significantly reduced, resulting in more spatially coherent change regions.

\begin{table}[!h]
\centering
\caption{Ablation study of different components within SCR. “JS” denotes Jensen-Shannon divergence, whereas “LogitDiff” represents logit-level discrepancy. Best performance is shown in \textbf{bold}.}
\label{tab:mod1_ablation}
\small
\renewcommand{\arraystretch}{1.2}
\setlength{\tabcolsep}{3pt}

\resizebox{\columnwidth}{!}{
\begin{tabular}{l|cc|cc|cc|cc}
\toprule
\textbf{Method} 
& \multicolumn{2}{c|}{\textbf{LEVIR-CD}} 
& \multicolumn{2}{c|}{\textbf{WHU-CD}} 
& \multicolumn{2}{c|}{\textbf{DSIFN}} 
& \multicolumn{2}{c}{\textbf{SECOND avg.}} \\

& \cellcolor{metricbg}$\mathrm{IoU}^{C}$ & \cellcolor{f1bg}$\mathrm{F}_{1}^{C}$ 
& \cellcolor{metricbg}$\mathrm{IoU}^{C}$ & \cellcolor{f1bg}$\mathrm{F}_{1}^{C}$ 
& \cellcolor{metricbg}$\mathrm{IoU}^{C}$ & \cellcolor{f1bg}$\mathrm{F}_{1}^{C}$ 
& \cellcolor{metricbg}$\mathrm{IoU}^{C}$ & \cellcolor{f1bg}$\mathrm{F}_{1}^{C}$ \\
\midrule

Baseline (XOR) 
% & 54.52 & 70.57 & 48.26 & 65.10 & 40.36 & 57.51 & 27.24 & 41.49 \\
%changed
& 53.28 & 69.52 & 48.26 & 65.10 & 40.36 & 57.51 & 27.24 & 41.49 \\

JS only 
% & 65.07 & 78.84 & 70.07 & 82.40 & 45.32 & 62.37 & 25.42 & 39.12 \\
%changed
& \textbf{65.05} & \textbf{78.83} & \textbf{70.07} & \textbf{82.40} & 45.32 & 62.37 & 25.42 & 39.12 \\

LogitDiff only 
% & 60.95 & 75.74 & 62.98 & 77.28 & 48.83 & 65.62 & 29.66 & 44.57 \\
%changed
& 63.92 & 77.99 & 62.98 & 77.28 & 48.83 & 65.62 & \textbf{29.66} & \textbf{44.57} \\

% CMC only 
% & -- & -- & -- & -- & -- & -- & 27.81 & 42.27 \\

Full SCR 
% & \textbf{61.67} & \textbf{76.29} 
%changed
& 64.38 & 78.33 
& 62.99 & 77.29
& \textbf{49.01} & \textbf{65.78} 
& 29.64 & 44.52 \\

\bottomrule
\end{tabular}
}
\end{table}

\subsubsection{Intra-module Ablation of SCR}
% Table~\ref{tab:mod1_ablation} presents the ablation of internal components within the SCR module.
% We have observed compelling cross-dataset behavior differences. 
% Specifically, using Jensen-Shannon divergence(JS) alone produces substantial improvements on all BCD datasets, suggesting it is highly effective for identifying whether semantic responses between bi-temporal images become inconsistent.
% However, the improvement becomes unstable on the SECOND, where the performance drops below the baseline. We attribute this phenomenon to the fact that JS divergence mainly captures relative distribution differences. However, in multi-category scenarios, low-confidence predictions across several semantically similar classes may still produce noticeable distribution shifts, leading to ambiguous or noisy change responses.
%
% In contrast, applying Logit Difference(LogitDiff) explicitly models response intensity variation and therefore achieves better robustness on complex semantic scenarios such as SECOND, while also maintaining competitive performance on DSIFN. This indicates that magnitude-aware discrepancy modeling is important for suppressing minor semantic fluctuations and highlighting meaningful category transitions.
% Overall, the full SCR achieves the most stable performance across all datasets. This indicates that jointly modeling distribution-level shifts and response variations provides a more reliable characterization of semantic change, consistent with our design hypothesis.
%changed
Table~\ref{tab:mod1_ablation} presents the ablation of internal components within SCR. Using JS divergence alone brings strong improvements on BCD datasets, especially on LEVIR-CD where $\mathrm{F}_{1}^{C}$ increases from $69.52\%$ to $78.83\%$. This indicates that distribution-level divergence is effective for identifying candidate changes with substantial semantic redistribution.
However, JS alone is less stable on SECOND, where multiple semantically similar categories may produce distribution shifts even without reliable target activation. In contrast, LogitDiff captures response-level magnitude changes and achieves stronger performance on SECOND, suggesting that response variation is important under multi-class semantic competition.
The full SCR combines these two cues and provides a more balanced semantic assessment across datasets. Although JS alone performs best on some BCD settings, combining distribution-level divergence and response-level variation improves robustness when semantic ambiguity and category competition become stronger.

\begin{table}[!h]
\centering
\caption{Ablation study of BCR. “Pixel-level BCR” retains only pixels with high boundary-aware confidence, while “Region-aware BCR” further validates connected regions based on interior high-confidence support. Best performance is shown in \textbf{bold}.}
\label{tab:mod2_ablation}
\small
\renewcommand{\arraystretch}{1.2}
\setlength{\tabcolsep}{3pt}

\resizebox{\columnwidth}{!}{
\begin{tabular}{l|cc|cc|cc|cc}
\toprule
\textbf{Method} 
& \multicolumn{2}{c|}{\textbf{LEVIR-CD}} 
& \multicolumn{2}{c|}{\textbf{WHU-CD}} 
& \multicolumn{2}{c|}{\textbf{DSIFN}} 
& \multicolumn{2}{c}{\textbf{SECOND avg.}} \\

& \cellcolor{metricbg}$\mathrm{IoU}^{C}$ & \cellcolor{f1bg}$\mathrm{F}_{1}^{C}$ 
& \cellcolor{metricbg}$\mathrm{IoU}^{C}$ & \cellcolor{f1bg}$\mathrm{F}_{1}^{C}$ 
& \cellcolor{metricbg}$\mathrm{IoU}^{C}$ & \cellcolor{f1bg}$\mathrm{F}_{1}^{C}$ 
& \cellcolor{metricbg}$\mathrm{IoU}^{C}$ & \cellcolor{f1bg}$\mathrm{F}_{1}^{C}$ \\
\midrule

Baseline (XOR) 
% & 54.52 & 70.57 & 48.26 & 65.10 & 40.36 & 57.51 & 27.24 & 41.49 \\
%changed
& 53.28 & 69.52 & 48.26 & 65.10 & 40.36 & 57.51 & 27.24 & 41.49 \\

Pixel-level BCR 
% & 52.84 & 69.15 & 70.40 & 82.63 & 40.63 & 57.78 & 26.68 & 41.10 \\
%changed
& 49.72 & 66.41 & 70.40 & 82.63 & 40.63 & 57.78 & 26.68 & 41.10 \\

Region-aware BCR 
% & \textbf{65.55} & \textbf{79.19} 
%changed
& \textbf{70.79} & \textbf{82.90} 
& \textbf{82.31} & \textbf{90.30} 
& \textbf{43.14} & \textbf{60.27} 
& \textbf{30.29} & \textbf{45.04} \\

\bottomrule
\end{tabular}
}
\end{table}

\subsubsection{Intra-module Ablation of BCR}

Table~\ref{tab:mod2_ablation} evaluates different spatial reliability strategies within BCR. Pixel-level BCR retains only pixels with high boundary-aware confidence. This is effective on WHU-CD, where $\mathrm{F}_{1}^{C}$ increases from $65.10\%$ to $82.63\%$, suggesting that boundary-induced artifacts are a major error source in sparse building scenes.
However, directly thresholding boundary-aware confidence can remove valid pixels near object boundaries. On LEVIR-CD, Pixel-level BCR decreases $\mathrm{F}_{1}^{C}$ from $69.52\%$ to $66.41\%$, showing that pixel-wise spatial filtering may fragment candidate regions when changes are dense or structurally irregular.
Region-aware BCR addresses this issue by validating connected candidate regions according to interior support instead of requiring every pixel to be individually reliable. This strategy improves performance across all datasets, reaching $82.90\%$ $\mathrm{F}_{1}^{C}$ on LEVIR-CD and $90.30\%$ on WHU-CD, which supports the use of region-level spatial assessment for filtering boundary-dominated candidates while preserving coherent changed regions.

\section{Conclusion}
In this paper, we propose ReA-OVCD, an efficient training-free framework for dense open-vocabulary change detection. ReA-OVCD first obtains dense semantic predictions for all queried categories through a shared open-vocabulary parsing process, which preserves category competition under the full vocabulary while avoiding repeated class-wise inference. Instead of directly treating pixel-wise semantic label discrepancies as final changes, it uses them to flexibly generate candidate change regions and then filters the unreliable ones through semantic-spatial reliability assessment.
Specifically, SCR retains discrete label discrepancies as flexible candidate-generation cues, but validates each candidate using distribution-level divergence and response-level variation, thereby suppressing label flips that lack sufficient semantic evidence. BCR further complements SCR from the spatial perspective by moving the decision from isolated pixels to connected regions. It checks whether candidate regions contain sufficient interior support and removes boundary-dominated responses caused by segmentation uncertainty and slight geometric misalignment.
Extensive experiments on LEVIR-CD, WHU-CD, DSIFN, and SECOND demonstrate that ReA-OVCD improves the reliability of pixel-level OVCD and achieves strong performance against recent OVCD methods with superior computational efficiency. Overall, ReA-OVCD represents a shift from direct pixel-wise semantic comparison to reliability-guided candidate validation, preserving fine-grained localization flexibility while providing a unified and efficient training-free pathway for both BCD and SCD.

% \section{Acknowledgments}

\bibliographystyle{ieeetr}

\bibliography{ref}

\end{document}